\documentclass[letterpaper, 10 pt, conference]{ieeeconf}
\usepackage[T1]{fontenc}
\usepackage{cite}
\usepackage{amssymb,amsfonts}
\usepackage{blindtext}
\makeatletter
\let\NAT@parse\undefined
\makeatother
\usepackage{hyperref}
\usepackage{graphicx}
\usepackage{textcomp}
\usepackage{mathtools}
\usepackage{changes}
\usepackage{subcaption}
\usepackage{algorithm}
\usepackage{xcolor}
\usepackage{color, soul}
\usepackage{amsmath}
\usepackage{booktabs}
\usepackage{multirow}
\usepackage{xstring}
\usepackage{hhline}
\usepackage{kotex}
\usepackage{siunitx}
\usepackage{comment}
\usepackage{physics}
\usepackage{tikz}
\usepackage{cleveref}
\usepackage{lipsum}
\usepackage{bm}
\usepackage{siunitx}
\usepackage{array}
\usepackage{eqparbox}
\usepackage{caption}
\usepackage[noend]{algorithmic}

\usepackage{etoolbox}  
\makeatletter
\patchcmd{\algorithmic}{\addtolength{\ALC@tlm}{\leftmargin} }{\addtolength{\ALC@tlm}{\leftmargin}}{}{}
\makeatother

\crefformat{section}{\S#2#1#3} 
\crefformat{subsection}{\S#2#1#3}
\crefformat{subsubsection}{\S#2#1#3}

\floatname{algorithm}{Algorithm}

\definecolor{rv}{RGB}{0, 0, 0}

\sethlcolor{lightgray}

\makeatletter
\DeclareRobustCommand{\iscircle}{\mathord{\mathpalette\is@circle\relax}}
\newcommand\is@circle[2]{%
  \begingroup
  \sbox\z@{\raisebox{\depth}{$\m@th#1\bigcirc$}}%
  \sbox\tw@{$#1\square$}%
  \resizebox{!}{\ht\tw@}{\usebox{\z@}}%
  \endgroup
}
\makeatother

\IEEEoverridecommandlockouts                              

\usepackage{amsmath} 
\usepackage{caption}
\newcommand{\rom}[1]{\uppercase\expandafter{\romannumeral #1\relax}}

\title{\LARGE \bf

TRAVEL: Traversable Ground and Above-Ground Object Segmentation Using Graph Representation of 3D LiDAR Scans}

\author{Minho Oh$^{1,\dag}$, Euigon Jung$^{1,\dag}$, Hyungtae Lim$^{1}$, Wonho Song$^{1}$, Sumin Hu$^{1}$, Eungchang Mason Lee$^{1}$\\ Junghee Park$^{2}$, Jaekyung Kim$^{2}$, Jangwoo Lee$^{2}$, and Hyun Myung$^{1,*}$, \textit{Senior Member, IEEE}
\thanks{$^*$Corresponding author: Hyun Myung}
\thanks{$^\dag$These authors contributed equally.}
\thanks{
$^1$The authors are with the School of Electrical Engineering and KI-AI at Korea Advanced Institute of Science and Technology (KAIST), Daejeon, 34141, Republic of Korea. {\tt\small \{minho.oh, egjung94, shapelim, swh4613, 2minus1, eungchang\_mason, hmyung\}@kaist.ac.kr} \hfill \break 
\indent $^{2}$The authors are with LIG Nex1 Co. Ltd., Seongnam 13488, Republic of Korea. {\tt\footnotesize \{junghee.park, jkkim0211, jangwoo.lee2\}@lignex1.com} \hfill \break
\indent \textcolor{rv}{This work was supported by grants from LIG Nex1 Co. Ltd. and BK21 FOUR.} \hfill \break
}
}

\begin{document}

\captionsetup[figure]{labelformat={default},labelsep=period,name={fig.}}

\maketitle
\thispagestyle{empty}
\pagestyle{empty}

\begin{abstract}
Perception of traversable regions and objects of interest from a 3D point cloud is one of the critical tasks in autonomous navigation. A ground vehicle needs to look for traversable terrains that are explorable by wheels. Then, to make safe navigation decisions, the segmentation of objects positioned on those terrains has to be followed up. However, over-segmentation and under-segmentation can negatively influence such navigation decisions. To that end, we propose \textit{TRAVEL}, which performs traversable ground detection and object clustering simultaneously using the graph representation of a 3D point cloud.
To segment the traversable ground, a point cloud is encoded into \textcolor{rv}{a graph structure, \textit{tri-grid field}, which treats each tri-grid as a node}. Then, the traversable regions are searched and redefined by examining local convexity and concavity of edges that connect nodes.
On the other hand, our above-ground object segmentation employs a graph structure by representing a group of horizontally neighboring 3D points in a spherical-projection space as a node and vertical/horizontal relationship between nodes as an edge. Fully leveraging the node-edge structure, the above-ground segmentation ensures real-time operation and mitigates over-segmentation.
Through experiments using simulations, urban scenes, and our own datasets, we have demonstrated that our proposed traversable ground segmentation algorithm outperforms other state-of-the-art methods in terms of the conventional metrics and that our newly proposed evaluation metrics are meaningful for assessing the above-ground segmentation. We will make the code and our own dataset available to public at
\href{https://github.com/url-kaist/TRAVEL}{\texttt{https://github.com/url-kaist/TRAVEL}}.
\end{abstract}

\begin{keywords}
Traversable ground segmentation; Object segmentation; Graph search; LiDAR; Autonomous navigation\end{keywords}

\vspace{-0.2cm}
\section{Introduction} \label{sec:intro}
\vspace{-0.2cm}

In recent years, there has been an increasing demand to perceive and represent surroundings in the robotics field. For autonomous navigation, robust object segmentation is required to identify meaningful objects---possibly to track and even avoid them in the subsequent autonomous tasks or utilize them in localization \cite{pan2021mulls, sung2021if} or mapping \cite{arora2021mapping, yoon2019mapless}. In this paper, we explicitly explore segmentation of a point cloud captured by a 3D LiDAR sensor. A typical point cloud segmentation infers a class label of each data point, which is mostly tackled by learning-based approaches. Unfortunately, most of these approaches are supervised with ground-truth labels and computationally costly to process an immense number of data on a mobile GPU. These downsides have hindered a mobile robot platform from adopting the learning approaches into navigation in an unknown environment.

\begin{figure}[t]
    \captionsetup{font=footnotesize}
	\centering
    \includegraphics[width=\columnwidth]{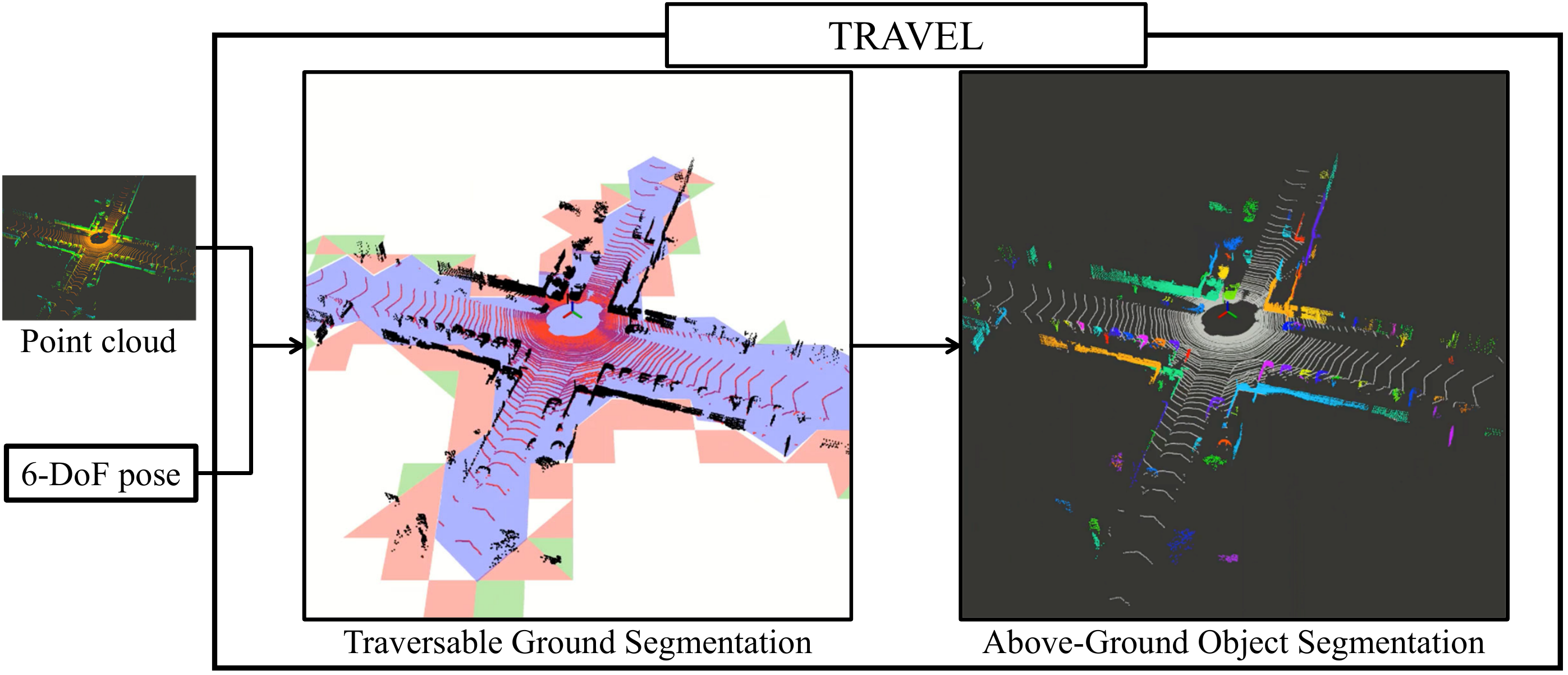}
    \caption{Overview of TRAVEL. TRAVEL segments a point cloud in two steps: traversable ground and above-ground object segmentation.}
    \label{fig:travel_overview}
    \vspace{-0.6cm}
\end{figure}
The focus of our work is primarily oriented towards safe navigation that is not subject to a specific environment. Rather than assigning class labels (e.g. vehicles, pedestrians, poles, etc.) to all data points, we detect a traversable area and spatially distinguish meaningful objects. To that end, we propose an efficient segmentation algorithm, named TRAVEL, that leverages a node-edge graph structure. TRAVEL consists of tri-grid field (TGF)-based traversable ground segmentation by terrain modeling and above-ground object segmentation by clustering, as shown in Fig.~\ref{fig:travel_overview}. We evaluate our proposed algorithm on CARLA \cite{dosovitskiy17}, Semantic KITTI dataset \cite{behley2019semantickitti}, and our own rough terrain dataset, to underscore the following contributions:

\begin{itemize}
    \item To the best of our knowledge, this research is the first to introduce TGF for ground segmentation. By leveraging TGF and geometrical discrepancy, our algorithm can effectively detect the traversability of terrains based on the proposed breadth-first search.
    \item We separately cluster object points horizontally and vertically with the following improvements:
        \begin{itemize}
            \item We introduce a concept of \textit{skipped linkage} and \textit{circular linkage} to effectively handle over-segmentation.
            \item Our binary search approach reduces the complexity of finding neighboring nodes to $\mathit{O}(N\log(N))$, where $N$ is the total number of nodes in a ring.
        \end{itemize}
    \item We propose novel metrics, namely \textit{over-segmentation entropy} and \textit{under-segmentation entropy}. These metrics measure the distribution and uncertainty of labels that correspond to an object, being comprehensive and meaningful measures to assess segmentation performance.
\end{itemize}

The rest of the letter is organized as follows: Section~\ref{sec:related_works} provides an overview of related works. Section~\ref{sec:tri-tzone} explains the proposed traversable ground segmentation algorithm; Section~\ref{sec:object_segmentation} delineates the procedure of the above-ground segmentation; Section~\ref{sec:experiment_setting} describes the experiments and novel evaluation metrics; Section~\ref{sec:result_and_discussion} discusses the experimental results; and, finally, Section~\ref{sec:conclusion} summarizes our contributions and explores future works.

\vspace{-0.3cm}

\section{Related Works}\label{sec:related_works}
\vspace{-0.2cm}

\subsection{Ground Segmentation}
\vspace{-0.1cm}

To overcome the under-segmentation problem, a real-time ground segmentation that is robust to varying environments should come prior to object segmentation. The previous lines of research base themselves on RANSAC \cite{fischler1981ransac, narksri2018slope} or ground plane fitting approaches using the principal component analysis (PCA) \cite{zermas2017fast, lim21patchwork}. Among them, Narksri \textit{et al.}~\cite{narksri2018slope} and Lim \textit{et al.}~\cite{lim21patchwork} proposed the slope-robust methods based on the ego-centric scan representation. On the other hand, Xue \textit{et al.}~\cite{xue2021lidar} proposed an algorithm that examines density of scan points based on the grid-based scan representation. All of the above methods have solely focused on ground segmentation itself and have not examined the traversability, discerning an area on which a vehicle can stand and travel safely. Moosmann \textit{et al.}~\cite{moosmann2009segmentation} employed a point-wise graph search method to split the ground and vertical points considering continuity by local convexity. However, traversability is still not considered, and real-time operation is not guaranteed due to point-wise graph-search. To address this issue, our traversable ground segmentation explores the traversability considering all three attributes: slope, level, and continuity of a terrain.

\subsection{Object Segmentation}
\vspace{-0.1cm}

Given the immense resources required to classify massive number of remaining scan points after the ground segmentation, above-ground object segmentation particularly focuses on real-time feasibility while keeping up reasonable accuracy. 
Bogoslavskyi \textit{et al.}~\cite{bogo} employed a range image and calculated the difference in reflection angles of adjacent points to differentiate objects.
Zermas \textit{et al.}~\cite{zermas2017fast} proposed a novel method of using the ring structure of a 3D point cloud captured by a 3D LiDAR sensor, clustering points horizontally in a ring and vertically between rings in two steps. 
Burger \textit{et al.}~\cite{mesh} proposed a mesh structure using loss functions that consist of cluster densities, slopes, distances, and angles; however, its three steps of horizontal, vertical, and fusion updates entail high computation complexity. 
Yang \textit{et al.}~\cite{node} represented a set of horizontally neighboring points as a node and created a set graph. However, Yang \textit{et al.}~\cite{node} did not consider the interaction between index-wise non-adjoining nodes, even when, due to occlusion, they can be possible candidates to be clustered geometrically. 
Establishing deficient interplay between scan points, the above segmentation methods are prone to over-segmentation, especially caused by occlusion, and under-segmentation, especially caused by ground segmentation failure. 
In addition, although the prior works measured computation times and qualitatively evaluated their performance, they lacked appropriate quantitative metrics to scrutinize over-segmentation and under-segmentation.
\vspace{-0.1cm}

\subsection{Learning-based Segmentation}

\vspace{-0.1cm}
Several works have tackled the segmentation task with learning-based methods. The research by Zhang \textit{et al.}~\cite{zhang2020instance} directly identified objects from a point cloud. Wong \textit{et al.}~\cite{wong2020identifying} proposed an instance segmentation method to recognize both known and unknown instance objects. Although these learning-based instance segmentation methods can accurately extract objects, they cannot recognize the ground or walls that are also important for autonomous navigation. Unlike the instance segmentation, others proposed the semantic segmentation method that can label the point-wise class~\cite{milioto2019rangenet++,xu2021rpvnet,peng2022mass}. These methods extract the ground without considering  traversability. Also, since the extracted objects do not have instance labels, some separate post-processes (e.g. clustering) are necessary to be applied to autonomous navigation. Paigwar \textit{et al.}~\cite{paigwar2020gndnet} proposed the semantic segmentation with only two classes; ground and non-ground. It can process the ground estimation in real-time by considering the slope of terrains. Although learning-based methods work well in the learned environment, the performance cannot be guaranteed under the condition encountered for the first time\textcolor{rv}{~\cite{wong2020identifying}}.
\section{Traversable Ground Segmentation}\label{sec:tri-tzone}
\vspace{-0.1cm}
TGF-based traversable ground segmentation, the first step of TRAVEL, mainly consists of three parts: node-wise terrain modeling on TGF, breadth-first traversable graph search (B-TGS), and traversable ground modeling and segmentation. 
\vspace{-0.2cm}
\subsection{Pre-processing for Traversability}
\vspace{-0.1cm}
\begin{figure}[b!]
    \captionsetup{font=footnotesize}
	\vspace{-0.4cm}
	\centering
    \begin{subfigure}[t]{0.45\columnwidth}
        \centering
        \includegraphics[width=\textwidth]{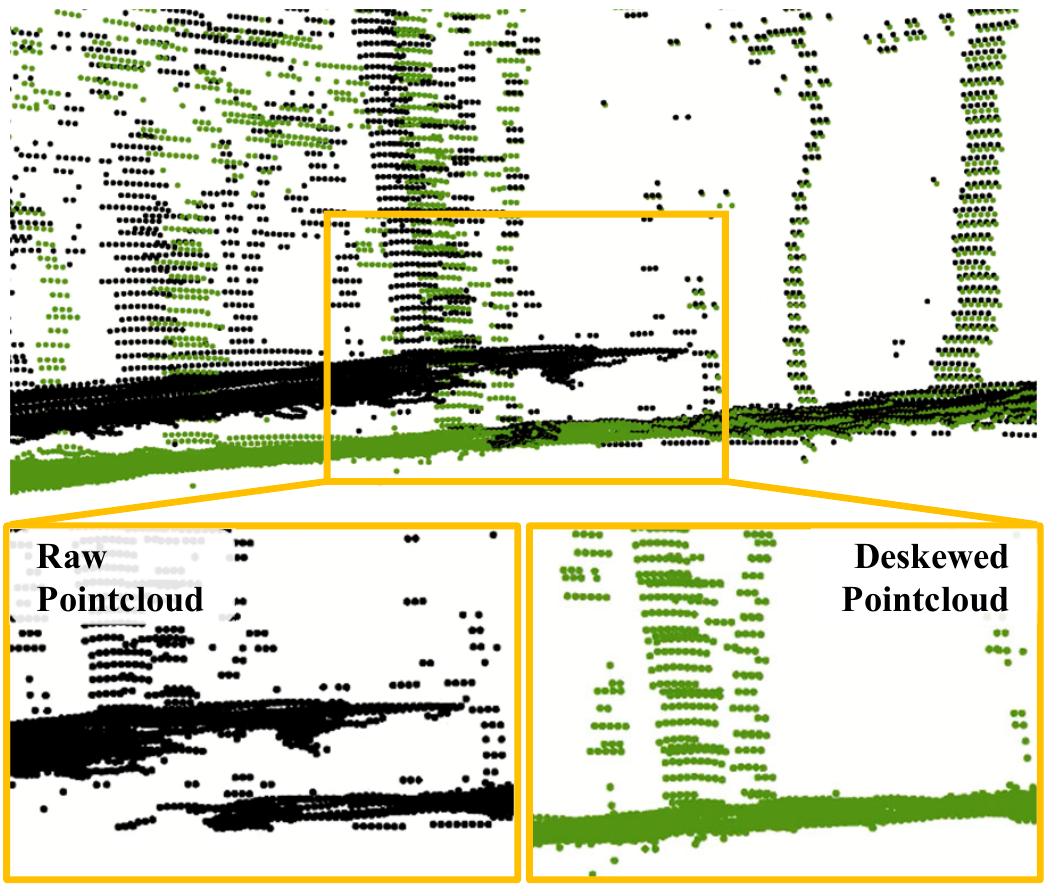}
        \caption{}
    \end{subfigure}
    \begin{subfigure}[t]{0.45\columnwidth}
        \centering
        \includegraphics[width=\textwidth]{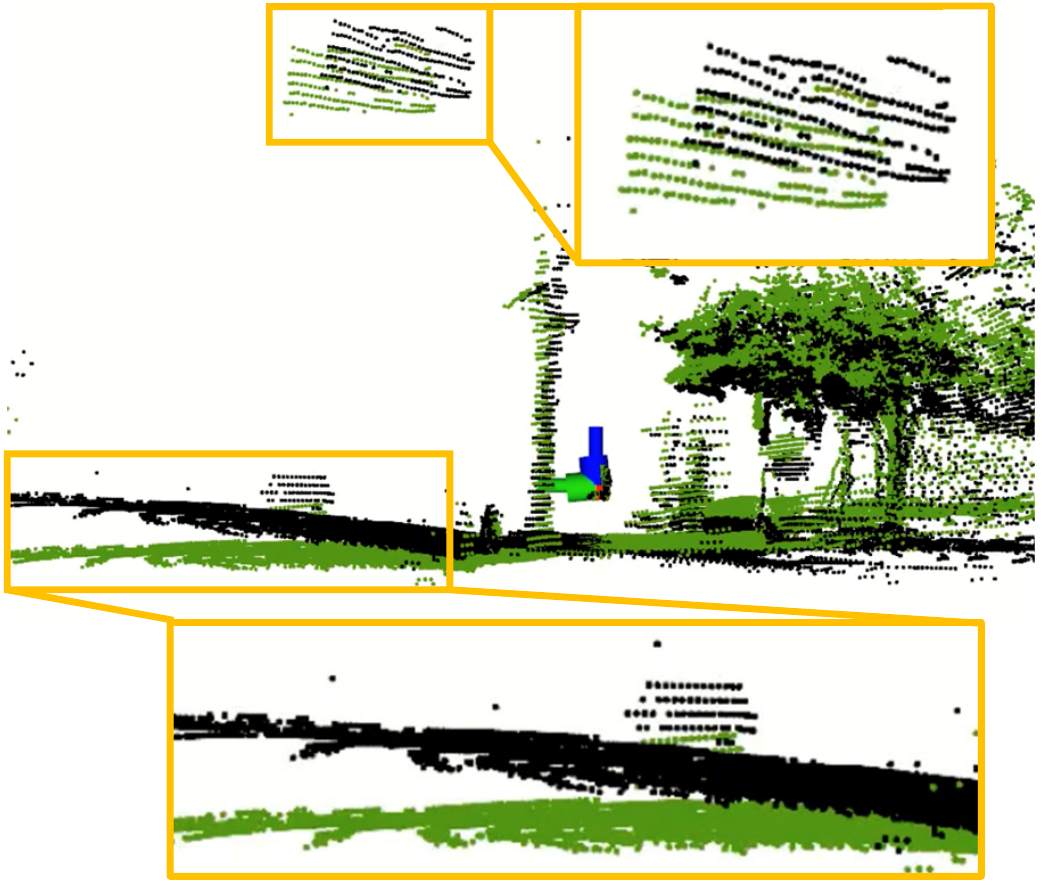} 
        \caption{}
    \end{subfigure}
    \vspace{-0.2cm}
    \caption{Effects by (a) point cloud deskewing and (b) attitude alignment. The black and green points represent point clouds before and after applying each compensation, respectively. (Best viewed in color)}
    \label{fig:pre_processing}
    \vspace{-0.3cm}
\end{figure}

The pre-processing step is essential for boosting segmentation performance especially in bumpy terrains. As shown in Fig.~\ref{fig:pre_processing}(a), a point cloud can be skewed by the motion of a mobile platform \cite{sung2021if, shan2020lio}. To deskew a point cloud, we use an IMU pre-integration approach to fuse LiDAR and IMU data tightly. Also, we adopt a 6-DoF pose to alleviate the segmentation problem caused by a tilted platform, as shown in Fig.~\ref{fig:pre_processing}(b). The raw point cloud is rotated to the upright position by using the pitch and roll angles calculated from the pose. After these compensations, we can accurately estimate the terrain properties in the world coordinate\cite{moosmann2009segmentation}.
\vspace{-0.1cm}
\subsection{Node-wise Terrain Model on TGF}
\vspace{-0.1cm}

\begin{figure}[t!]
    \captionsetup{font=footnotesize}
    \centering
    \begin{subfigure}[t]{0.40\columnwidth}
        \includegraphics[width=\textwidth]{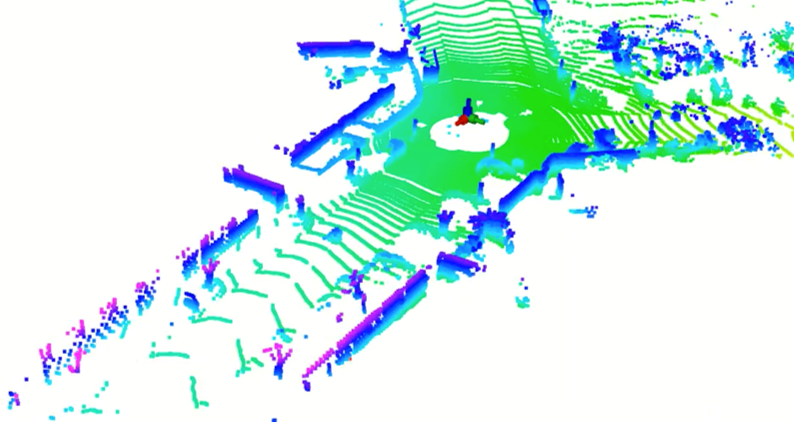}
        \caption{}
    \end{subfigure}
    \begin{subfigure}[t]{0.40\columnwidth}
        \includegraphics[width=\textwidth]{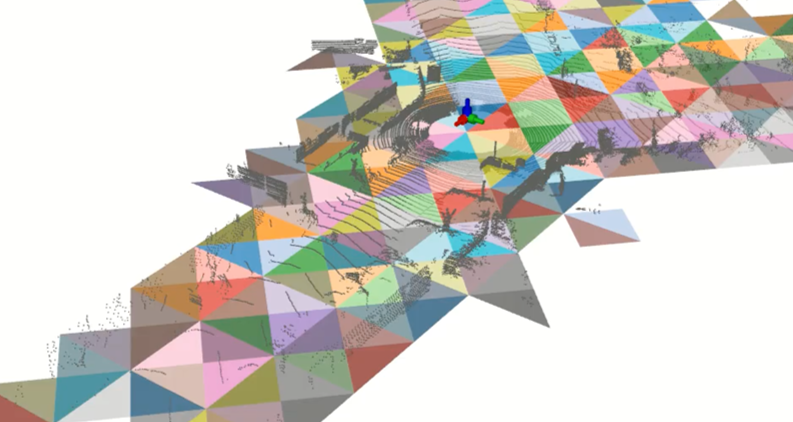}
        \caption{}
    \end{subfigure}
    \caption{(a) An example of a 3D point cloud. (b) Representation of Tri-Grid Field (\textit{TGF}). A point cloud is encoded into TGF, and each triangular node contains the points based on their $\mathit{xy}$-coordinates.}
    \label{fig:tri_grid_embedding}
    \vspace{-0.3cm}
\end{figure}

\begin{figure}[t!]
    \captionsetup{font=footnotesize}
    \centering
    \begin{subfigure}[t]{0.324\columnwidth}
        \centering
        \includegraphics[width=\textwidth]{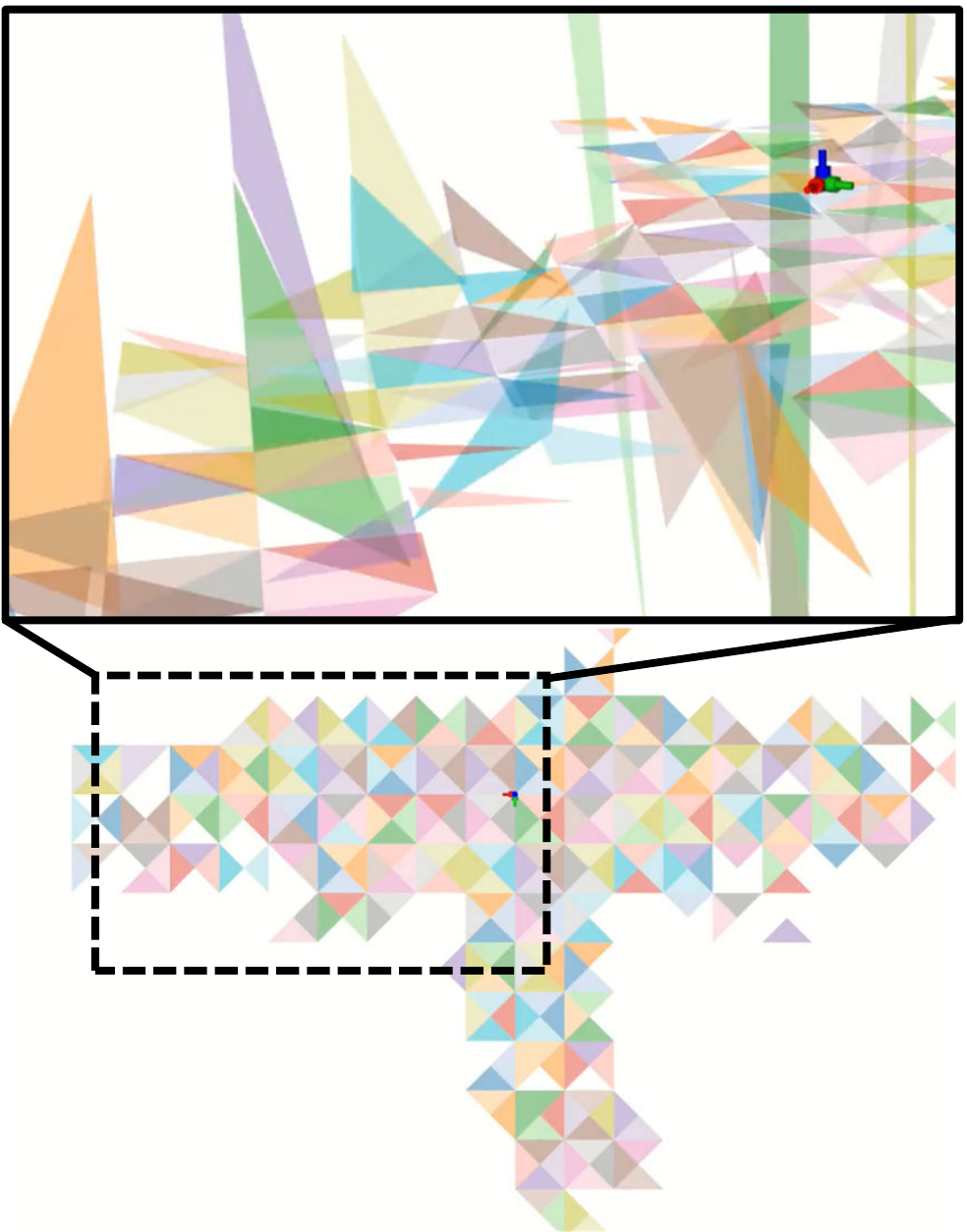}
        \caption{}
    \end{subfigure}
    \begin{subfigure}[t]{0.324\columnwidth}
        \centering
        \includegraphics[width=\textwidth]{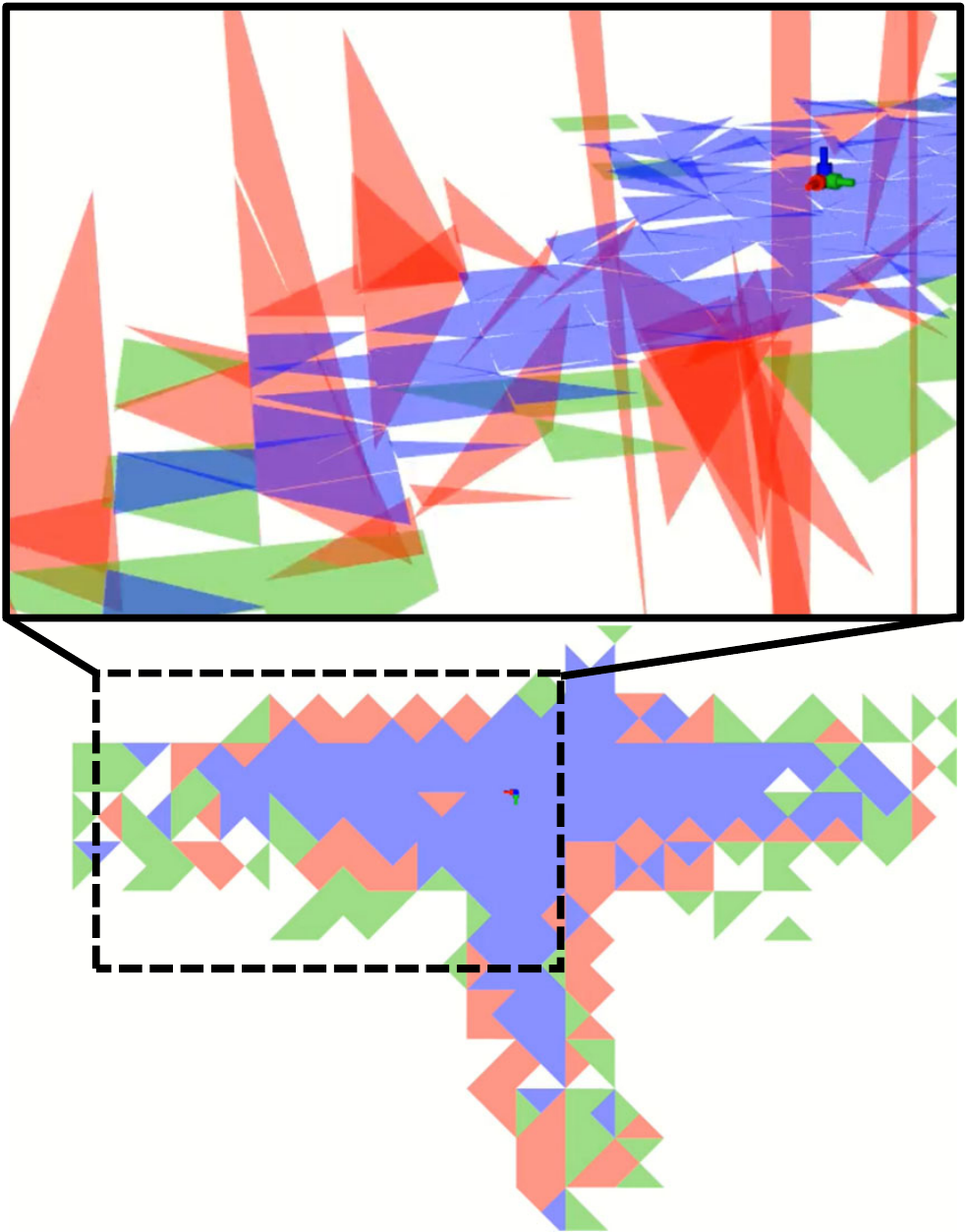}
        \caption{}
    \end{subfigure}
    \begin{subfigure}[t]{0.324\columnwidth}
        \centering
        \includegraphics[width=\textwidth]{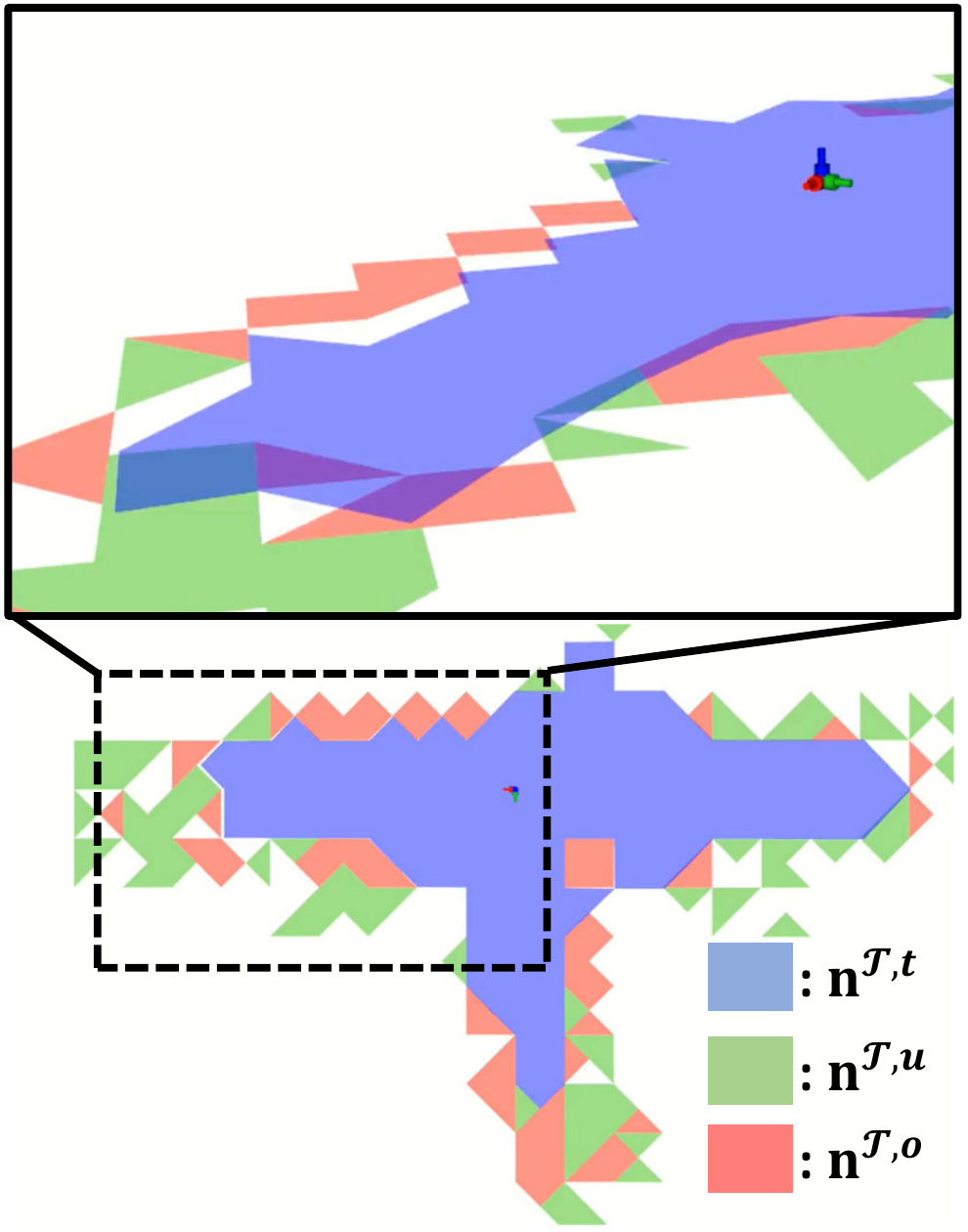}
        \caption{}
    \end{subfigure}
    \vspace{-0.15cm}    
    \caption{Changes in \textit{TGF} at each process step for traversable ground segmentation.
            (a) The terrain in each node is modeled independently.
            (b) The nodes are classified into terrain (blue), obstacle (red), and unknown (green) nodes by their models.
            (c) The traversable terrains are modeled based on the nodes searched by \textit{B-TGS}, and the planar models of non-traversable nodes are rejected. (Best viewed in color)
            }
    \label{fig:TGF_flow}
\vspace{-0.7cm}
\end{figure}

Firstly, as shown in Fig.~\ref{fig:tri_grid_embedding}, a 3D point cloud is encoded into our graph representation, TGF, which is the pre-built field on the $\mathit{xy}$-coordinates with a constant resolution, ${r^\mathcal{T}}$. And each triangular-shaped node $\mathbf{n}^{\mathcal{T}}_{i}$ contains a corresponding partial point cloud,~$\Bbb{P}_i$.
TGF consists of a set of nodes $\mathbf{N}^{\mathcal{T}}=\{\mathbf{n}^{\mathcal{T}}_{i}|i\in\mathcal{N}\}$ and a set of edges $\mathbf{E}^{\mathcal{T}}=\{\mathbf{e}^{\mathcal{T}}_{ij}|i,j\in\mathcal{N}\}$, where $\mathbf{e}^{\mathcal{T}}_{ij}$ connects $\mathbf{n}^{\mathcal{T}}_{i}$ and $\mathbf{n}^{\mathcal{T}}_{j}$, and $\mathcal{N}$ is a set of node indices. 

Accordingly, the PCA-based ground plane fitting approach estimates a planar model $\mathbf{P}_i$, initial ground points, and descending ordered eigenvalues $\lambda_{k\in\{1,2,3\}}$ for $\Bbb{P}_i$ in each $\mathbf{n}^{\mathcal{T}}_{i}$, where $\mathbf{P}_i$ consists of a normalized surface normal vector $\mathbf{s}_i\in \mathbb{R}^3$, i.e. $\norm{\mathbf{s}_i}=1$, and plane coefficient $d_i$~\cite{zermas2017fast, lim21patchwork}. A mean point $\mathbf{m}_i\in \mathbb{R}^3$, which is obtained by averaging the initial ground points among $\Bbb{P}_i$, and weight $\mathit{w}^{\mathcal{T}}$($\mathbf{n}^{\mathcal{T}}_i$) for scoring the traversability of a corresponding tri-grid are also included in $\mathbf{n}^{\mathcal{T}}_{i}$. They are denoted as follows:
\begin{equation}
\begin{split}
    \mathbf{P}_{i}^{\mathsf{T}} 
    \begin{bmatrix}
        \mathbf{m}_{i}\\
        1
    \end{bmatrix} 
    &= 
    \begin{bmatrix}
    \mathbf{s}_{i}^{\mathsf{T}} & d_{i}
    \end{bmatrix}
    \begin{bmatrix}
        \mathbf{m}_{i}\\
        1
    \end{bmatrix}
    = 0,
    \\
    w^{\mathcal{T}}(\mathbf{n}^{\mathcal{T}}_{i}) &= (\textit{cohesion}+\textit{planarity})/\textit{linearity}\\
                    &= \lambda_{2,i}\cdot(\lambda_{1,i}+\lambda_{2,i})/(\lambda_{1,i}\cdot\lambda_{3,i})
\end{split}
\end{equation}
where $\textit{cohesion}=\lambda_{1}/\lambda_{3}$, $\textit{planarity}=\lambda_{2}/\lambda_{3}$, and $\textit{linearity}=\lambda_{1}/\lambda_{2}$ are the characteristic coefficients derived from the distribution of points, inspired by the study of \textcolor{rv}{Weinmann \textit{et al.}~\cite{weinmann2015semantic}.} Each node can be expressed as a tri-grid with the corresponding $\mathbf{P}$ as in Fig.~\ref{fig:TGF_flow}(a). Then, $\mathbf{n}^{\mathcal{T}}$ is classified into three types according to the inclination threshold parameter $\theta^{\mathcal{T}}$ and the number of points, $\sigma^{\mathcal{T}}$: terrain node $\mathbf{n}^{\mathcal{T},t}$, obstacle node $\mathbf{n}^{\mathcal{T},o}$, and unknown node $\mathbf{n}^{\mathcal{T},u}$. 
A node that is less inclined than $\theta^{\mathcal{T}}$, more inclined than $\theta^{\mathcal{T}}$, or has fewer points than $\sigma^{\mathcal{T}}$, is classified into $\mathbf{n}^{\mathcal{T},t}$, $\mathbf{n}^{\mathcal{T},o}$, and $\mathbf{n}^{\mathcal{T},u}$, respectively. Only terrain nodes are passed to the next step.

\vspace{-0.1cm}
\subsection{Breadth-first Traversable Graph Search (B-TGS)}
\vspace{-0.1cm}

\begin{figure}[t!]
    \captionsetup{font=footnotesize}
    \centering
    \begin{subfigure}[t]{0.60\columnwidth}
        \centering
        \includegraphics[width=\textwidth]{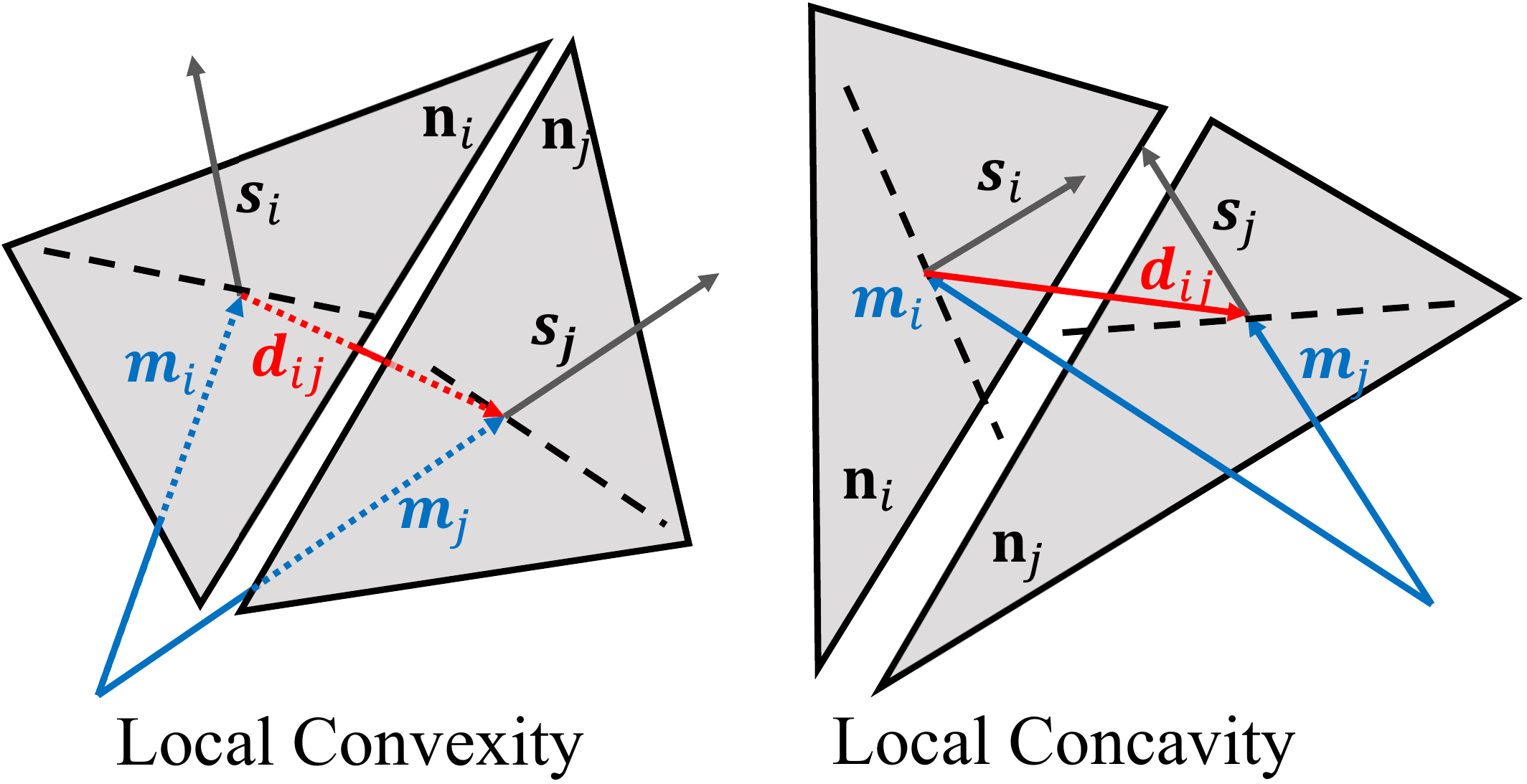}
        \caption{}
    \end{subfigure}
    \begin{subfigure}[t]{0.30\columnwidth}
        \centering
        \includegraphics[width=\textwidth]{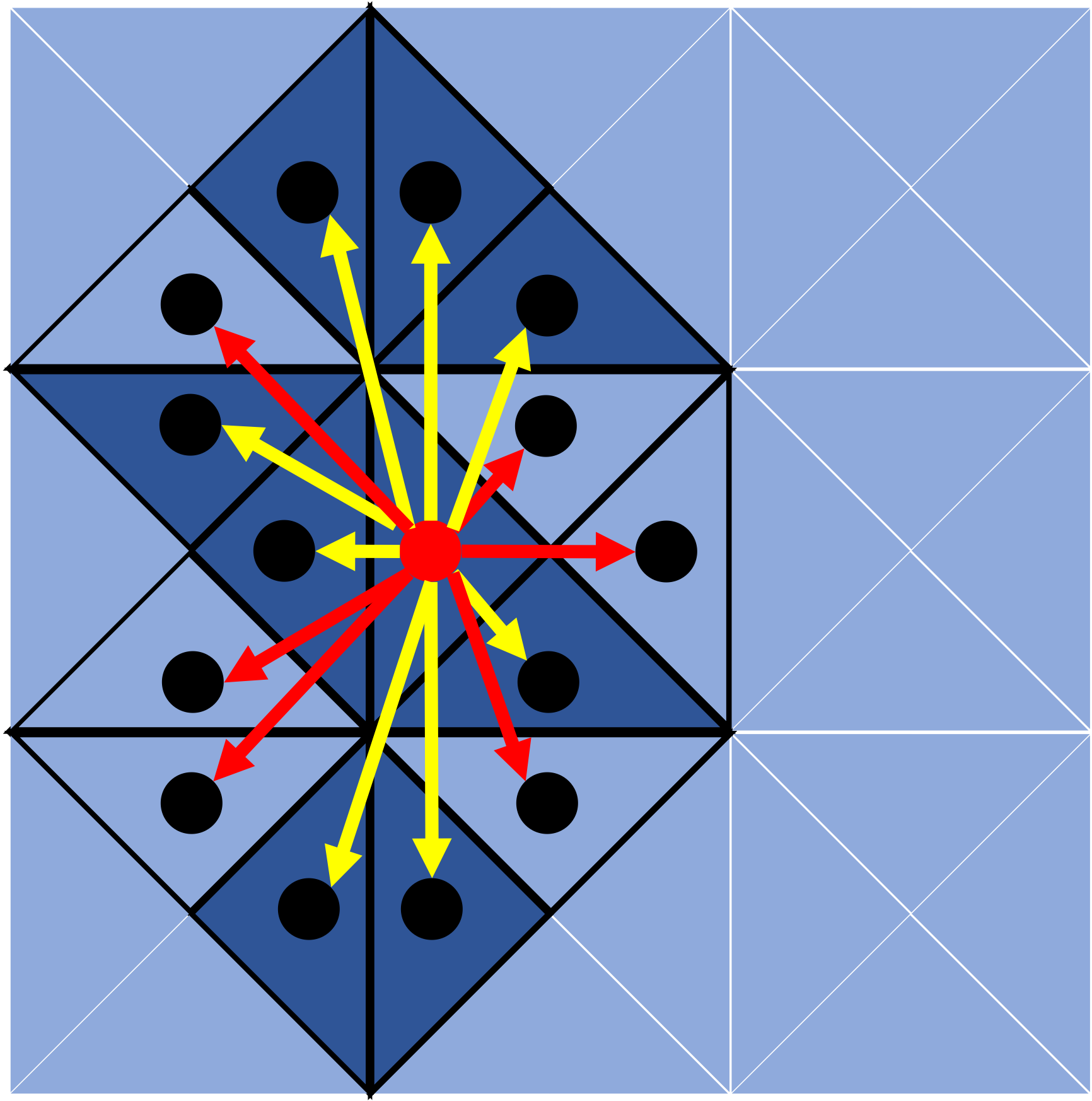}
        \caption{}
    \end{subfigure}
    \caption{
    (a) Local convexity/concavity of an edge considering the geometric relationship between two nodes.
    (b) An example of the first step in B-TGS. The neighboring nodes (black dots) are connected by the edges (arrows) from the seed (red dot). By examining local convexity/concavity of the edges, the final traversable edges and nodes are decided and illustrated in yellow arrows and dark blue grids, respectively.}
    \label{fig:local_convexity_concavity}
    \vspace{-0.3cm}
\end{figure}

\begin{figure}[t!]
    \captionsetup{font=footnotesize}
    \centering
    \begin{subfigure}[t]{0.43\columnwidth}
        \centering
        \includegraphics[width=\textwidth]{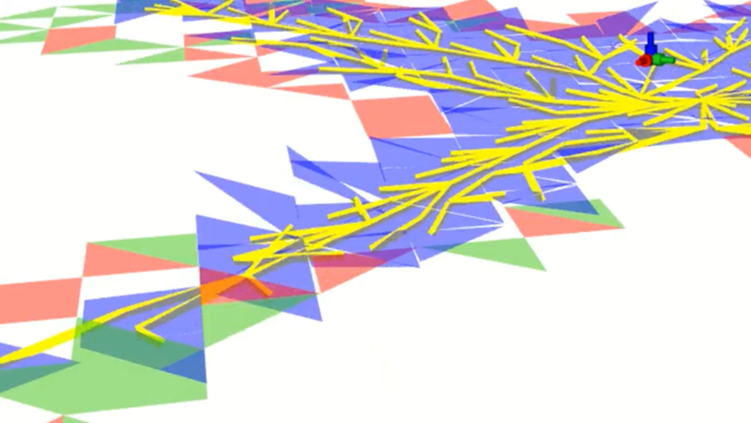}
        \caption{}
    \end{subfigure}
    \begin{subfigure}[t]{0.43\columnwidth}
        \centering
        \includegraphics[width=\textwidth]{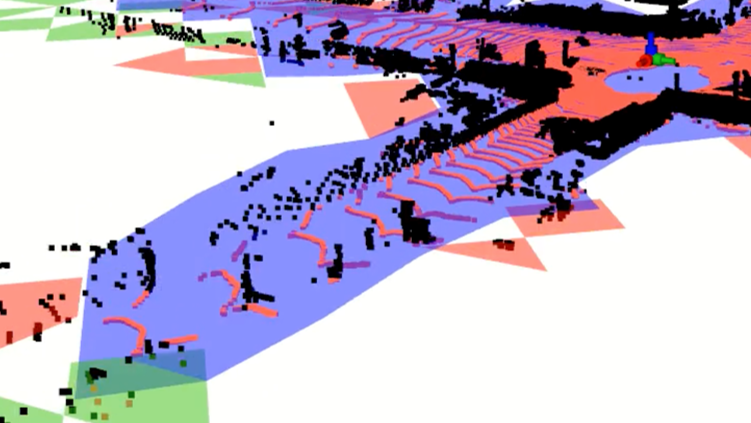}
        \caption{}
    \end{subfigure}
    \vspace{-0.15cm}
    \caption{(a) Breadth-first Traversable Graph Search (\textit{B-TGS}) result, where yellow lines define the traversable edges connecting the traversable terrain nodes. (b) Traversable Terrain Model Fitting (\textit{TTMF}) and segmentation result based on traversable nodes and edges. The red points represent the ground and the black ones indicate the obstacles above the traversable areas.}
    \label{fig:btgs_result}
    \vspace{-0.7cm}
\end{figure}

To search for a set of traversable nodes $\mathbf{T}^{\mathcal{T}}$ among a set of $\mathbf{n}^{\mathcal{T},t}$, we adopt a breadth-first search approach into TGF, namely \textit{B-TGS}.
At first, $\mathbf{n}^{\mathcal{T},t}$ with the highest weight and the closest to the sensor is selected as a seed $\mathbf{n}^{\mathcal{T},t}_{i_s}$, and the adjacent nodes of the seed are considered as the neighbors, as illustrated in Fig.~\ref{fig:local_convexity_concavity}(b).
Then, to determine the traversable node $\mathbf{n}^{\mathcal{T},t}_{i_n}$ among the neighbors, we calculate the traversability of $\mathbf{e}^{\mathcal{T}}_{i_{s}i_{n}}$ that is the geometric relationship between $\mathbf{n}^{\mathcal{T}}_{i_{s}}$ and $\mathbf{n}^{\mathcal{T}}_{i_{n}}$.

Inspired by \cite{moosmann2009segmentation}, the traversability of $\mathbf{e}^{\mathcal{T}}_{ij}$, is determined by local convexity and concavity of itself, which are illustrated in Fig.~\ref{fig:local_convexity_concavity}(a). The following equation expresses whether an edge $\mathbf{e}^{\mathcal{T}}_{ij}$ has acceptable local convexity and concavity: 
\begin{equation}
lcc(\mathbf{e}^{\mathcal{T}}_{ij}) = \begin{cases} 
    true,  & \text{if } (\,|\mathbf{s}_i\cdot\mathbf{s}_j| > 1 - \sin(||\mathbf{d}_{ij}||\epsilon_2)\,) \\ 
    & \wedge\:(\,|\mathbf{s}_j\cdot\mathbf{d}_{ji}|<||\mathbf{d}_{ji}||\sin\epsilon_1\,)\\
    & \wedge\:(\,|\mathbf{s}_i\cdot\mathbf{d}_{ij}|<||\mathbf{d}_{ij}||\sin\epsilon_1\,)\\
    false, & \text{otherwise}
\end{cases}
\end{equation}
where \(\mathbf{d}_{ji}=\mathbf{m}_i-\mathbf{m}_j\) is the displacement vector between two nodes, \(\epsilon_1\) denotes the angle at which \(\mathbf{m}_i\) (or \(\mathbf{m}_j\)) may lie above \(\mathbf{P}_j\) (or \(\mathbf{P}_i\)), and \(\epsilon_2\) denotes the threshold angle for similarity. If $lcc(\mathbf{e}^{\mathcal{T}})$ between the seed and other node is \textit{true}, then the node belongs to $\mathbf{T}^{\mathcal{T}}$ and becomes another seed for the following search steps. B-TGS continues until there are no more neighboring nodes.  
As shown in Fig.~\ref{fig:btgs_result}(a), at the end of B-TGS, the traversable set of $\mathbf{e}^{\mathcal{T}}$ is stretched out by connecting the corresponding $\mathbf{n}^{\mathcal{T},t}\in\mathbf{T}^{\mathcal{T}}\subset\mathbf{N}^{\mathcal{T}}$.

\vspace{-0.1cm}
\subsection{TGF-wise Traversable Terrain Model Fitting (TTMF)}
\vspace{-0.1cm}
Finally, in traversable terrain model fitting (TTMF) process, by applying the weighted corner fitting to the triangular corners $\mathbf{c}^{\mathcal{T}}_{k\in\{1,2,3\},i}=(x_{\mathbf{c}^{\mathcal{T}}_{k,i}},y_{\mathbf{c}^{\mathcal{T}}_{k,i}},z_{\mathbf{c}^{\mathcal{T}}_{k,i}})$ of the nodes $\mathbf{n}^{\mathcal{T},t}_{i}\in\mathbf{T}^{\mathcal{T}}$, the overall traversable ground on TGF is refined from $\mathbf{P}$ to $\hat{\mathbf{P}}$ for each $\mathbf{n}^{\mathcal{T},t}$.
In the weighted corner fitting, all the corners are grouped into $\mathbf{C}^{\mathcal{T}}_{m\in\mathcal{M}}=\{\mathbf{c}^{\mathcal{T}}_{k,i,m}|x_{\mathbf{c}_m}=x_{\mathbf{c}^{\mathcal{T}}_{k,i,m}}, y_{\mathbf{c}_m}=y_{\mathbf{c}^{\mathcal{T}}_{k,i,m}},\text{for } ^{\forall}i\in\mathcal{N} \}$, where $\mathcal{M}$ is a set of grouped corner indices. Then, to make a node with a high weight have a more significant effect on the model of its neighboring nodes, the height of each corner in $\mathbf{C}^{\mathcal{T}}_{m}$ is updated to $\hat{z}_{\mathbf{c}_m}$, i.e., $\hat{\mathbf{c}}_{m}=(x_{\mathbf{c}_m},y_{\mathbf{c}_m},\hat{z}_{\mathbf{c}_m})$ through the following weighted average step:
\begin{equation} 
    \hat{z}_{\mathbf{c}_{m}}=\frac{\sum_{\mathbf{C}^{\mathcal{T}}_{m}}(z_{\mathbf{c}^{\mathcal{T}}_{k,i,m}}\cdot \mathit{w}^{\mathcal{T}}(\mathbf{n}^{\mathcal{T}}_{i})/||\mathbf{c}^{\mathcal{T}}_{k,i,m}-\mathbf{m}_i||_{xy})}{\sum_{\mathbf{C}^{\mathcal{T}}_{m}}(\mathit{w}^{\mathcal{T}}(\mathbf{n}^{\mathcal{T}}_{i})/||\mathbf{c}^{\mathcal{T}}_{k,i,m}-\mathbf{m}_i||_{xy})}.
\end{equation}
Furthermore, every $\mathbf{n}^{\mathcal{T}}\in\mathbf{N}^{\mathcal{T}}$ surrounded by three $\hat{\mathbf{c}}_{m}$, is corrected to a terrain node, $\hat{\mathbf{n}}^{\mathcal{T},t}$ with the following updated elements:
\begin{equation}
\begin{split}
    \hat{\mathbf{P}} = 
        \begin{bmatrix}
            \hat{\mathbf{s}} & \hat{d}
        \end{bmatrix}
    &,\:\: \hat{\mathbf{m}} = (\hat{\mathbf{c}}_{1}+\hat{\mathbf{c}}_{2}+\hat{\mathbf{c}}_{3})/3
    \\
    \hat{d} = - \hat{\mathbf{s}} \cdot \hat{\mathbf{m}}
    &,\:\:
    \hat{\mathbf{s}} =\frac{(\hat{\mathbf{c}}_{2}-\hat{\mathbf{c}}_{1})}{||\hat{\mathbf{c}}_{2}-\hat{\mathbf{c}}_{1}||} 
                      \cross 
                      \frac{(\hat{\mathbf{c}}_{3}-\hat{\mathbf{c}}_{1})}{||\hat{\mathbf{c}}_{3}-\hat{\mathbf{c}}_{1}||}.
\end{split}
\end{equation}
As shown in Fig.~\ref{fig:btgs_result}(b), from $\hat{\mathbf{P}}$ for each $\hat{\mathbf{n}}^{\mathcal{T},t}$, the point cloud is segmented as follows:
\begin{equation}
\begin{split}
        \texttt{label}(\mathbf{p}_{k})&=
        \begin{cases}
        \texttt{Terrain},& \text{if }\mathbf{p}_{k}\cdot\hat{\mathbf{s}}_i+\hat{d}_i < \epsilon_3
        \\
        \texttt{Obstacle},& \text{otherwise}
        \end{cases}
\end{split}
\end{equation}
for a point $\mathbf{p}_{k} \in \Bbb{P}_i$, where $\epsilon_3$ denotes the point-to-plane distance threshold.

\section{Above-Ground Object Segmentation}
\label{sec:object_segmentation}
The above-ground object segmentation first processes a spherical projection of a point cloud $\mathbb{P}$ that has points $\texttt{label}(\mathbf{p})=\texttt{Obstacle}$. Then, it executes the horizontal and vertical update iteratively on each row in the projection space for the efficient computation time. 
\vspace{-0.2cm}

\begin{figure}[ht!]
    \captionsetup{font=footnotesize}
    \centering
    \includegraphics[width=\columnwidth]{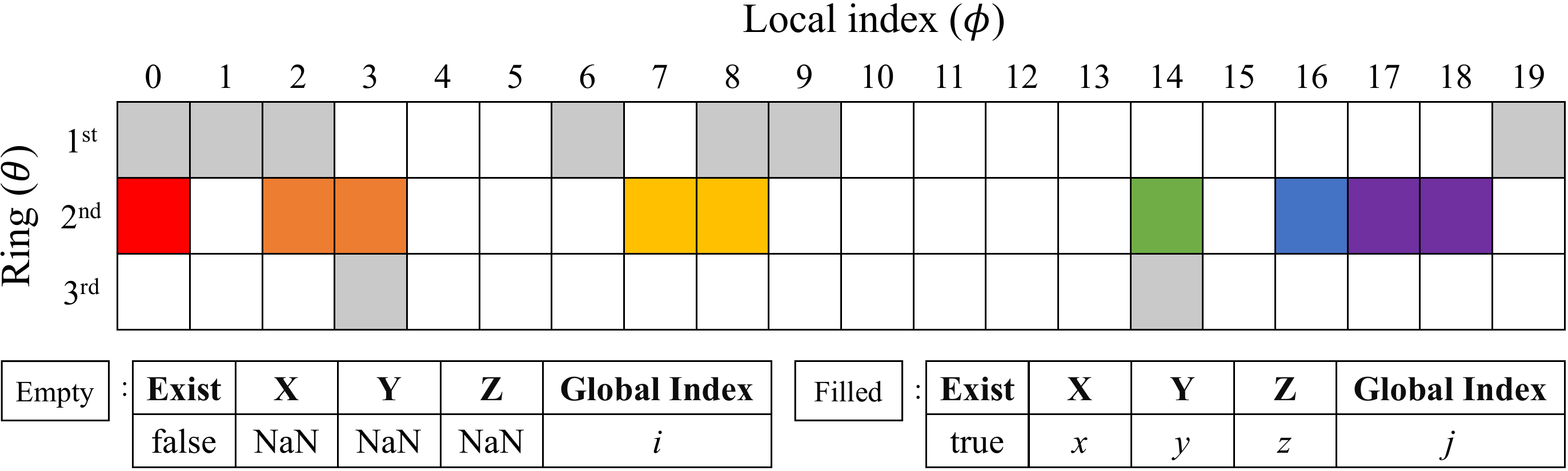}
    \vspace{-0.15cm}
    \caption{A single scan is projected to a spherical projection space by the azimuth ($\phi$) and elevation ($\theta$) angles of points. Each of the filled containers indicates a measured point. Note that the containers in the second ring are colored to better elucidate the concept of a node and an edge in Fig.~\ref{fig:node-edge}.}
    \label{fig:spherical}
    \vspace{-0.5cm}
\end{figure}
\vspace{-0.1cm}

\subsection{Spherical Projection}
\vspace{-0.1cm}
A spherical projection~\cite{milioto2019rangenet++} can be helpful in capturing spatial adjacency of points by their angles of reflection, making the upcoming horizontal and vertical update steps simpler. The spherical projection of a point cloud arranges points by their azimuth and elevation angles of reflection by mapping into an $\mathbb{R}^2$ space with width $w$ and height $h$. Fig.~\ref{fig:spherical} shows an example of spherically projected points. 

\subsection{Graph Representation}
Similar to \cite{node}, horizontally adjacent points that are separated by a distance below a threshold $T_{\text{horz}}$ form a \textit{node} $\mathbf{n}^{\mathcal{C}}=(\text{idx}_{s}, \text{idx}_{e}, \texttt{label})$, where $\text{idx}_{s}$, $\text{idx}_{e}$, and $\texttt{label}$ indicate its local index of a starting point, an ending point, and its label, respectively. Since points are organized in the horizontal order in terms of their azimuth angles, registering local start and end indices of a node can be useful for subsequent clustering. In brief, a set of directed edges $\mathbf{E}^{\mathcal{C}}=\{\mathbf{e}^{\mathcal{C}}_{\theta_{i},\theta_{j}}|i,j\in\mathcal{N}_{\theta}, \theta_{i}, \theta_{j}\in\mathbb{T}\}$, each of which connects two nodes, and a set of nodes $\mathbf{N}^{\mathcal{C}}=\{\mathbf{n}^{\mathcal{C}}_{\theta, k}|\theta\in\mathbb{T}, k\in\mathcal{N}_{\theta}  \}$ constitute our graph structure $\mathbf{G}^{\mathcal{C}}$ for clustering, where $\mathbb{T}=\{0,...,h-1\}$ and $\mathcal{N}_{\theta}$ is a list of node indices in a ring $\theta$.

\begin{figure}[t!]
    \captionsetup{font=footnotesize}
    \centering
    \begin{subfigure}[t]{0.49\columnwidth}
        \centering
        \includegraphics[width=\textwidth]{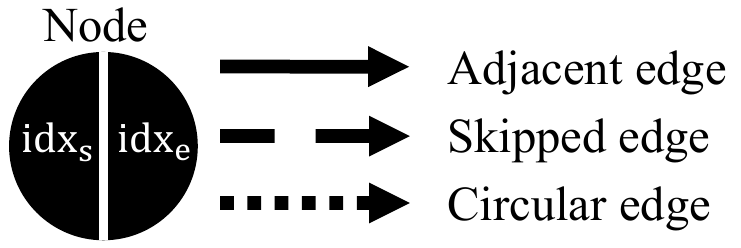}
    \end{subfigure}
    \begin{subfigure}[t]{0.49\columnwidth}
        \centering
        \includegraphics[width=\textwidth]{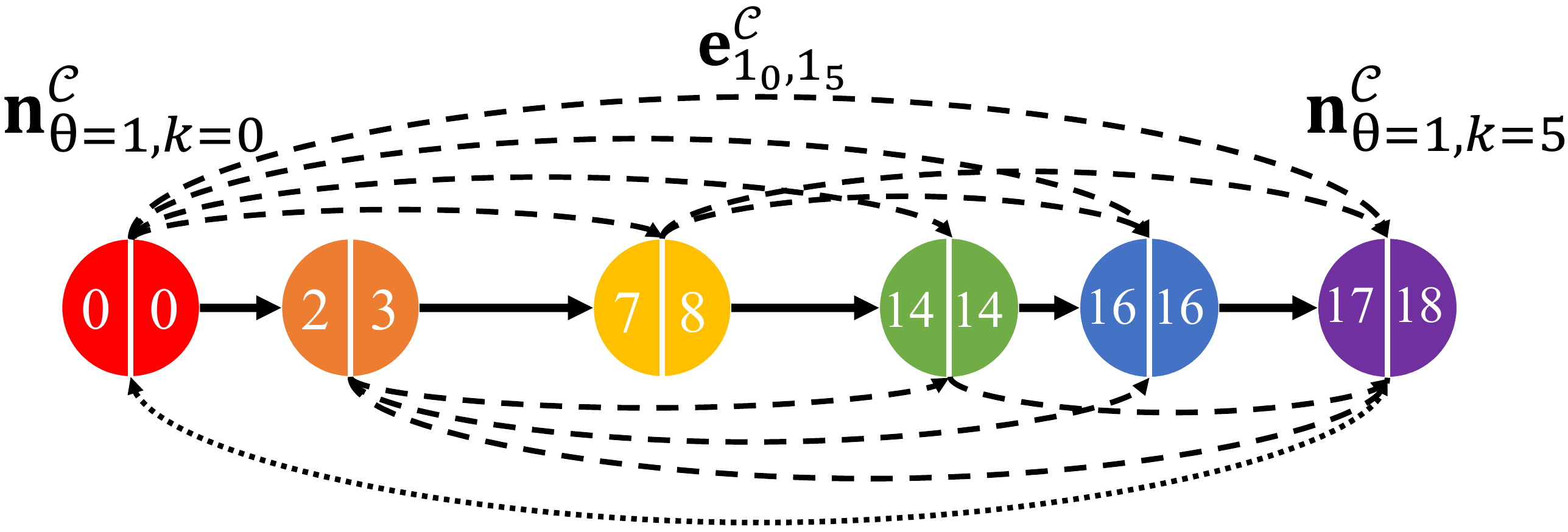}
        \caption{}
    \end{subfigure}
    \begin{subfigure}[t]{0.49\columnwidth}
        \centering
        \includegraphics[width=\textwidth]{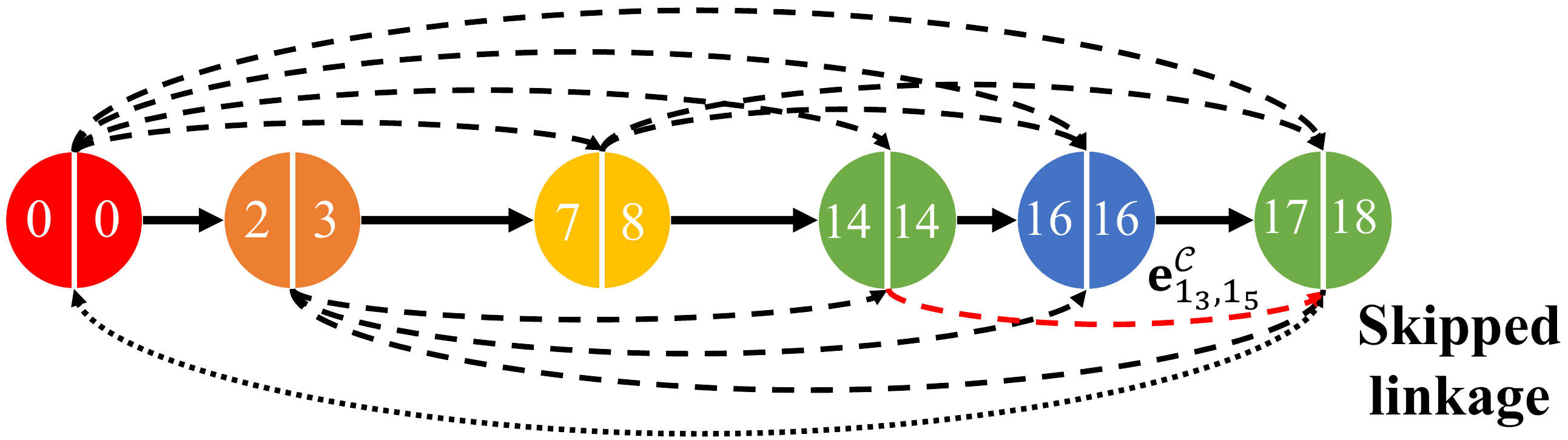}
        \caption{}  
    \end{subfigure}
    \begin{subfigure}[t]{0.49\columnwidth}
        \centering
        \includegraphics[width=\textwidth]{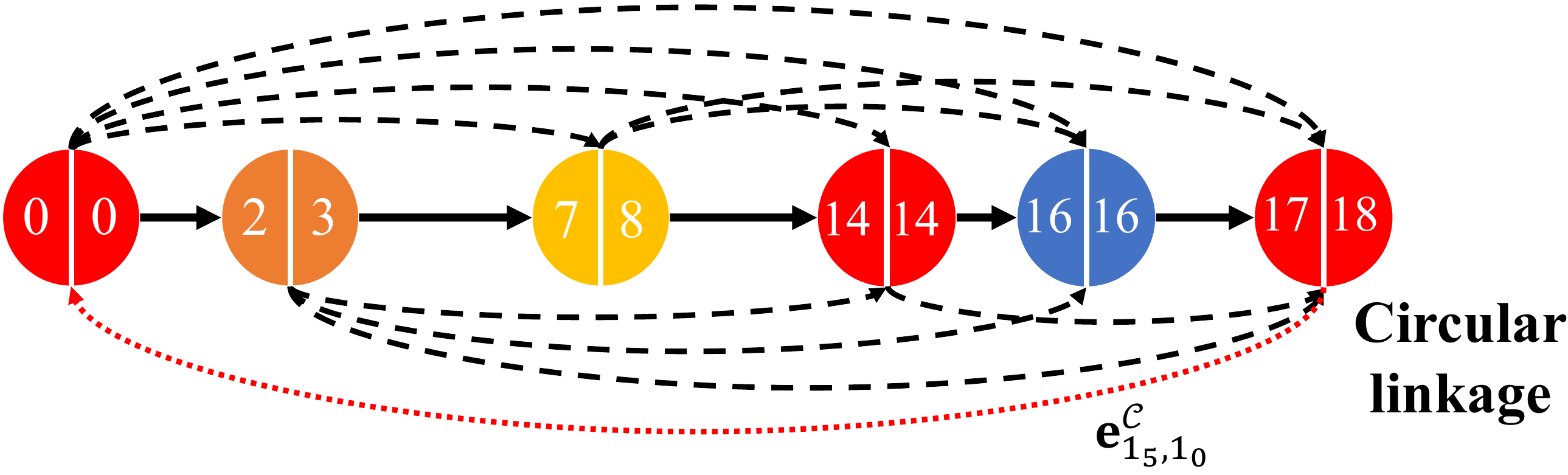}
        \caption{}  
    \end{subfigure}
    \vspace{-0.15cm}
    \caption{(a) Constructed from the second ring of the scan in Fig.~\ref{fig:spherical}, nodes are colored by its label and connected by directed edges. (b) The green and purple nodes belong to the same object but are separated by the blue node due to occlusion; two separated nodes pass the skipped linkage test and the label of the latter node is updated from purple to green, following the color of the former node. (c) The last and first nodes pass the circular linkage test, so the labels of the fourth and last nodes are updated to red, following the color of the first node. (Best viewed in color)}
    \label{fig:node-edge}
    \vspace{-0.3cm}
\end{figure}

\begin{figure}[t!]
    \captionsetup{font=footnotesize}
    \centering
    \begin{subfigure}[t]{0.49\columnwidth}
    \centering
        \includegraphics[width=\textwidth]{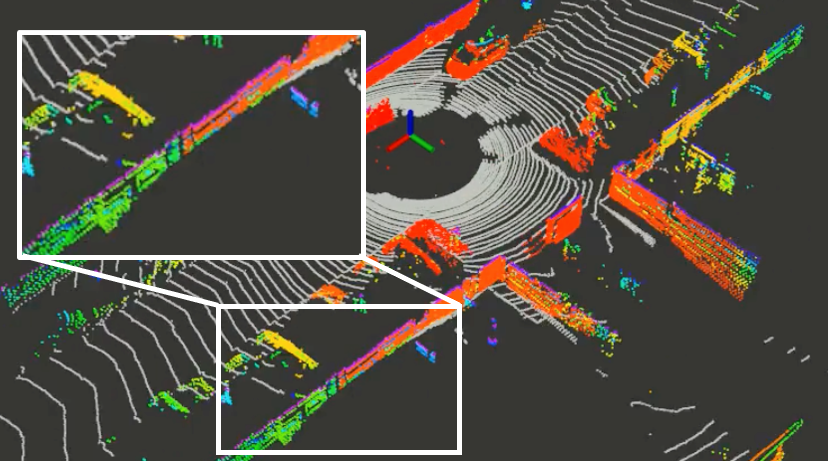}
        \caption{}
        \label{fig:wall_over}
    \end{subfigure}
        \begin{subfigure}[t]{0.49\columnwidth}
    \centering
        \includegraphics[width=\textwidth]{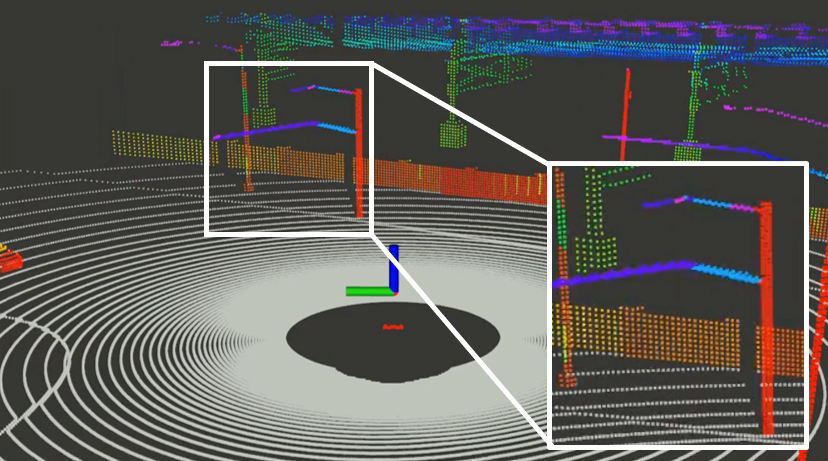}
        \caption{}
        \label{fig:pole_under}
    \end{subfigure}
    \vspace{-0.15cm}
    \caption{The same colored points represent one cluster. (a) The wall is over-segmentated as the objects in front of the wall prevents the LiDAR from sensing behind the structure. (b) The pole-like structure is over-segmented due to the lack of vertical clustering. (Best viewed in color)}
    \label{fig:seg_failure}
    \vspace{-0.4cm}
\end{figure}

\subsection{Horizontal Update}
The horizontal update step creates a node by clustering neighboring points within a horizontal merge threshold $T_{\text{horz}}$ along the direction of the local index. After the node-edge construction through the horizontal update, the second ring in Fig.~\ref{fig:spherical} can be compactly depicted as in Fig.~\ref{fig:node-edge}. 

By efficiently utilizing the compact representation, the circular linkage and skipped linkage tests can prevent over-segmentation due to occlusion. The circular linkage test checks whether the horizontal distance between the point in $\text{idx}_{e}$ of the last node $\mathbf{n}^{\mathcal{C}}_{\theta, n-1}$ and that in $\text{idx}_{s}$ of the first node $\mathbf{n}^{\mathcal{C}}_{\theta, 0}$ in the same ring $\theta$, falls below $T_{\text{horz}}$, where $n$ is the total number of nodes in $\theta$. This linkage check is necessary for preventing separation of points from a 3D LiDAR. The skipped linkage test merges non-neighboring nodes as long as these two nodes are located within $T_{\text{horz}}$. Likewise, the skipped linkage test plays its critical role under a frequently occurring situation as in Fig.~\ref{fig:seg_failure}(a).

\subsection{Vertical Update}\label{ssec:vert_update}
\begin{figure}[t!]
    \captionsetup{font=footnotesize}
    \centering
    \begin{subfigure}[t]{0.49\columnwidth}
        \includegraphics[width=0.5\textwidth]{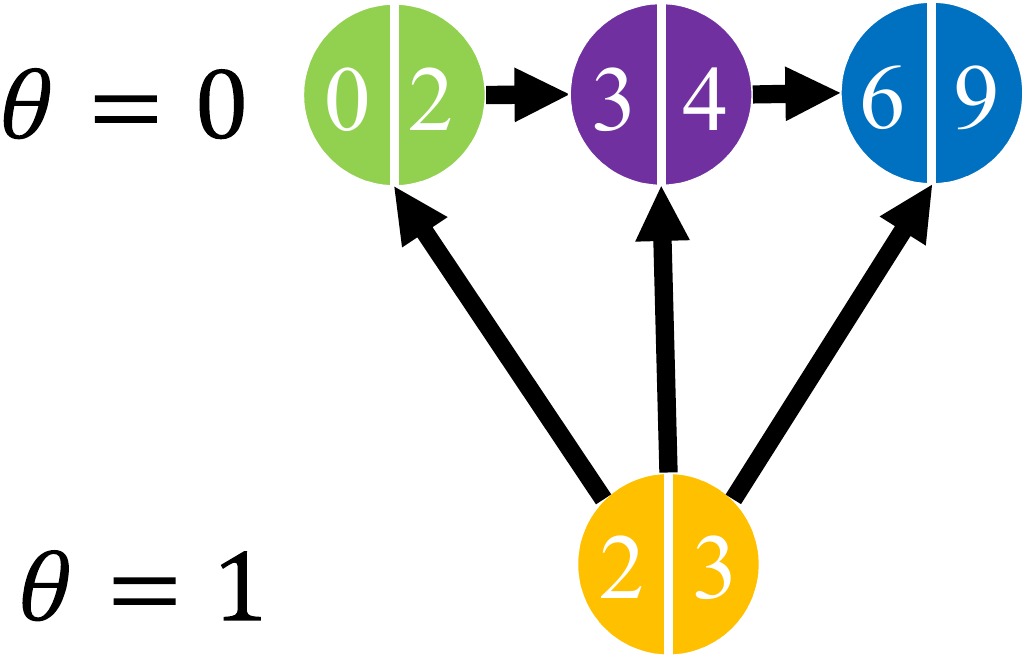}
    \end{subfigure}
    \begin{subfigure}[t]{0.49\columnwidth}
        \includegraphics[width=0.5\textwidth]{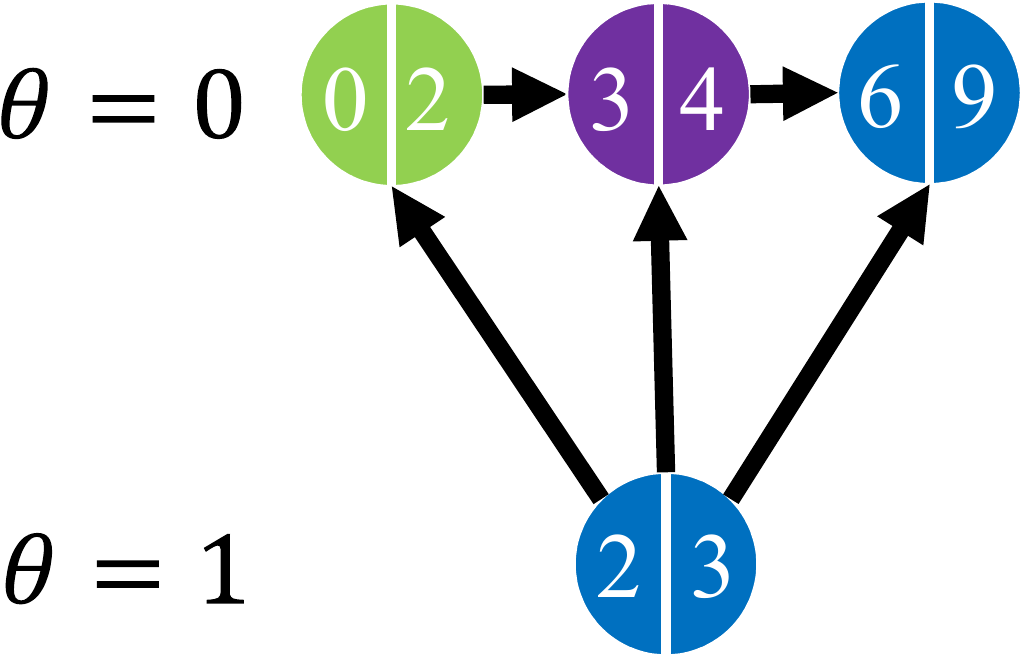}
    \end{subfigure}
    \begin{subfigure}[t]{0.49\columnwidth}
        \centering
        \includegraphics[width=0.9\textwidth]{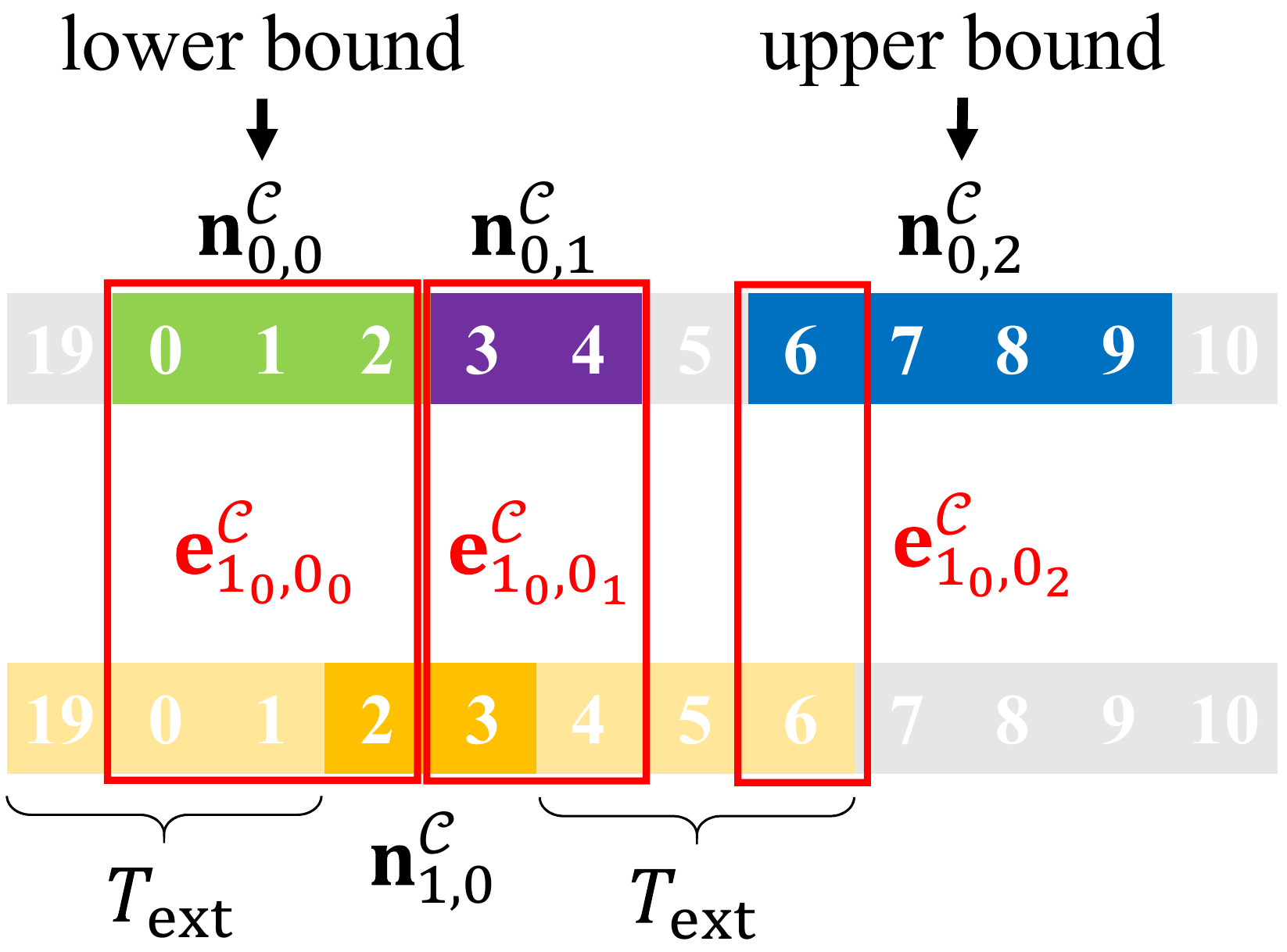}
        \caption{}
        \label{fig::vertical_update_a}
    \end{subfigure}
    \begin{subfigure}[t]{0.49\columnwidth}
        \centering
        \includegraphics[width=0.9\textwidth]{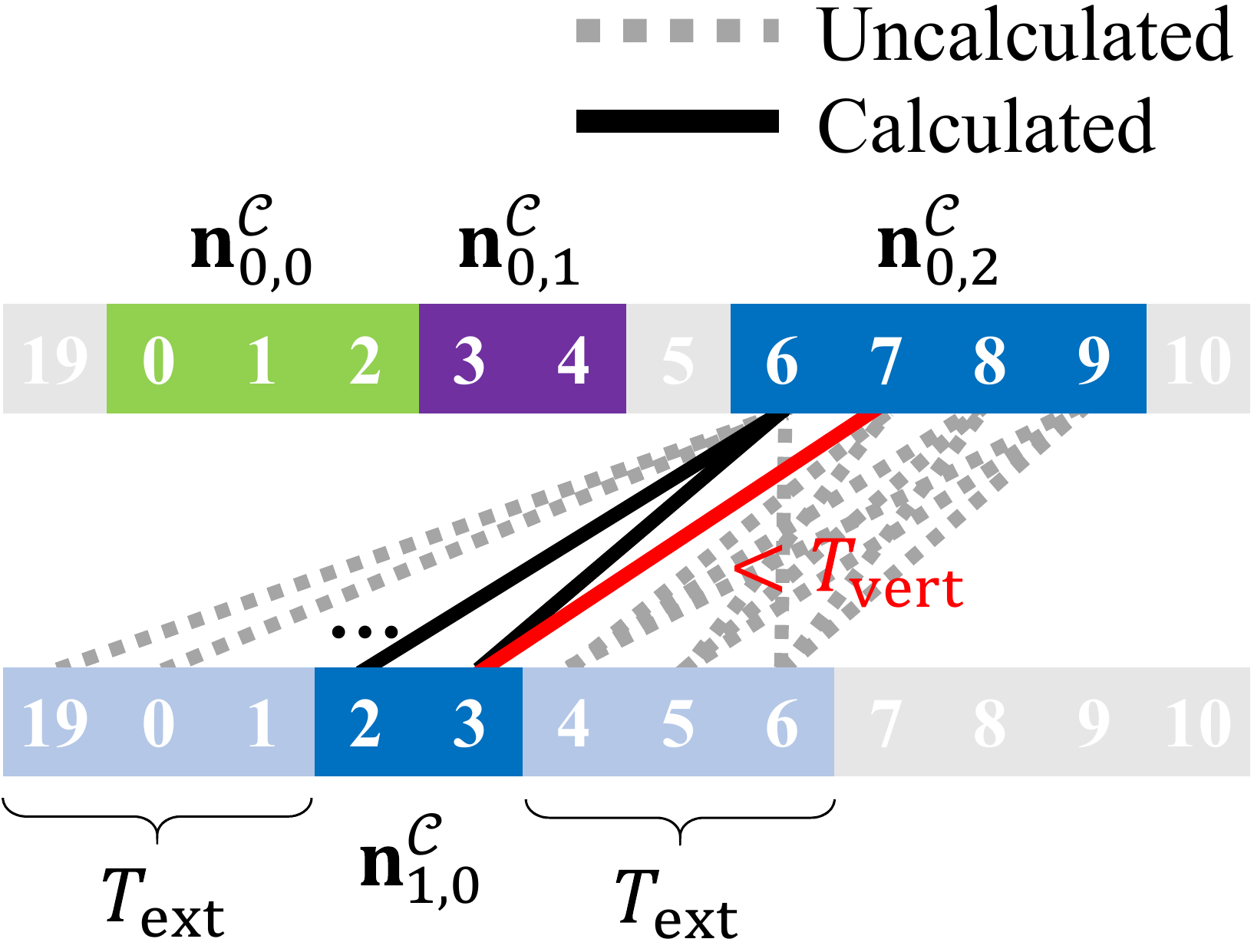}
        \caption{}
        \label{fig:vertical_update_b}
    \end{subfigure}
    \vspace{-0.15cm}
    \caption{An illustration of our vertical label update process. (a) Inter-ring edges. (b) Label update after computing the edge distances $D_{v}(\mathbf{e}^{\mathcal{C}}_{1_0,0_2})$.}
    \label{fig:vertical_update}
    \vspace{-0.7cm}
\end{figure}

Once a ring finishes the horizontal update, it goes through the vertical update; these two procedures are repeatedly executed on each ring. The chief novelty of our vertical update procedure lies in adopting binary search in finding overlaps and computing an edge distance $D_v$ between two nodes efficiently. As shown in Fig. \ref{fig:vertical_update}(a), \textcolor{rv}{the search space is first extended by the extension window size} $T_\text{ext}$ so that more possible edges are constructed, reducing chances of vertical separation as in Fig.~\ref{fig:seg_failure}(b). Then, at the previous ring, binary search is used to find the lower- and upper-bound nodes that have index-wise overlaps with the node $\mathbf{n}^{\mathcal{C}}_{1,0}$ at the current ring. This step can create inter-ring linkages from the current node to all nodes in between the lower- and upper-bounds. 

Fig. \ref{fig:vertical_update}(b) demonstrates how the vertical distance $D_v$ of an inter-ring edge is computed. In a naive way, $D_{v}(\mathbf{e}^{\mathcal{C}}_{1_0, 0_2})$ is defined by the distance between the points in $\mathbf{n}^{\mathcal{C}}_{1,0}$ and $\mathbf{n}^C_{0,2}$. However, the order of computing the distances becomes critical if a large number of points were to be clustered into a node. Since scan points that are collected at the same local index in multiple rings are more likely measured from one object, our algorithm starts computing the point distance from the overlapping index. In this way, our algorithm can retrench computational costs by saving a considerable number of redundant calculations.

\vspace{-0.1cm}
\section{Experiments} \label{sec:experiment_setting}
\vspace{-0.2cm}

\begin{table}[h]
    \vspace{-0.2cm}
    \captionsetup{font=footnotesize}
    \centering
    \caption{Parameter setting for TRAVEL. Units of $r^{\mathcal{T}}$, $\epsilon_{3}$, $T_{\text{horz}}$, and $T_{\text{vert}}$ are in unit of $m$, and $\theta^{\mathcal{T}}$ is in the unit of degree(\textdegree).}
    \setlength{\tabcolsep}{2.8pt}
    \setlength\extrarowheight{3.5pt}
    \begin{tabular}{l|c|c|c|c|c|c|c|c|c|c|c}
        \hline
        Param. & $r^{\mathcal{T}}$ & $\theta^{\mathcal{T}}$ & $\sigma^{\mathcal{T}}$ & $\epsilon_{1}$ & $\epsilon_{2}$ & $\epsilon_{3}$ & $T_{\text{horz}}$ & $T_{\text{skip}}$ & $T_{\text{ring}}$ & $T_{\text{vert}}$ & $T_{\text{ext}}$\\
        \hline
        Value & 8 & 30\textdegree & 0.1 & 0.03 & 0.1 & 0.1 & 0.3 & 10 & 5 & 0.5 & 100 \\
        \hline
    \end{tabular}
    \label{tab:parems}
    \vspace{-0.7cm}
\end{table}

\subsection{Dataset}
\vspace{-0.1cm}
To evaluate our proposed algorithm, we use the simulation, public urban scene dataset, and our own dataset. Both conventional and newly proposed evaluation metrics are used for quantitative evaluation. Table~\ref{tab:parems} shows the parameters that are used throughout the evaluation.

\subsubsection{CARLA Simulation}
In CARLA\cite{dosovitskiy17} simulation, the points in the scan have semantic labels. The points with the labels, such as \texttt{RoadLane}, \texttt{Road}, \texttt{SideWalk}, \texttt{Ground}, and \texttt{Terrain}, are considered to be the ground-truth terrain. Also, object identification, given by CARLA simulation, is used as ground-truth cluster label for the evaluation of above-ground object segmentation. This dataset is designed to test the algorithms in an urban environment with tunnels, slopes, buildings, and vehicles. 

\subsubsection{Semantic KITTI Dataset}
To evaluate the terrain segmentation performance of our proposed method against other ground segmentation algorithms, we experimented with the SemanticKITTI dataset\cite{behley2019semantickitti}, which presents an urban scene in real world. Note that the points that are labeled as \texttt{Lane marking}, \texttt{Road}, \texttt{Parking}, \texttt{Sidewalk}, \texttt{Other ground}, and \texttt{Terrain} are considered to be the ground-truth terrain points. \textcolor{rv}{The ground-truth cluster for each of the above-ground data points is decided by vanilla Euclidean clustering (EC).} This vanilla EC can cluster the points with the same class label and object identification, provided by Semantic KITTI dataset, as one object. Note that, for example, a tree in the dataset has points at its trunk labeled as \texttt{Trunk} and at its leaves as \texttt{Vegetation}, but our algorithm can only segment the tree as \textcolor{rv}{a single object}. Due to such limitation of the dataset, the evaluation of our algorithm will have some inevitable errors.

\subsubsection{Rough Terrain Dataset}
Unlike the other two datasets, our own rough terrain dataset introduces a challenging environment with bumpy terrains and low-lying obstacles. The dataset was acquired via a mobile robot that roams around in the forest and on sidewalks and roads. This robot platform, Husky from Clearpath Robotics, is equipped with a 3D LiDAR sensor of Ouster OS0-128 and an IMU sensor of Xsens MTI-300.

\vspace{-0.1cm}
\subsection{Evaluation Metrics}
\vspace{-0.1cm}
The traversable ground segmentation performance of TRAVEL is evaluated and compared with other ground segmentation algorithms through the conventional metrics: \textit{precision} (P), \textit{recall} (R), \textit{accuracy}, and \textit{F$_{1}$-score}. On the other hand, the above-ground object segmentation is evaluated by our newly proposed metrics: over-segmentation entropy~(OSE) and under-segmentation entropy~(USE).

\begin{figure}[t!]
    \captionsetup{font=footnotesize}
    \centering
    \includegraphics[width=0.75\columnwidth]{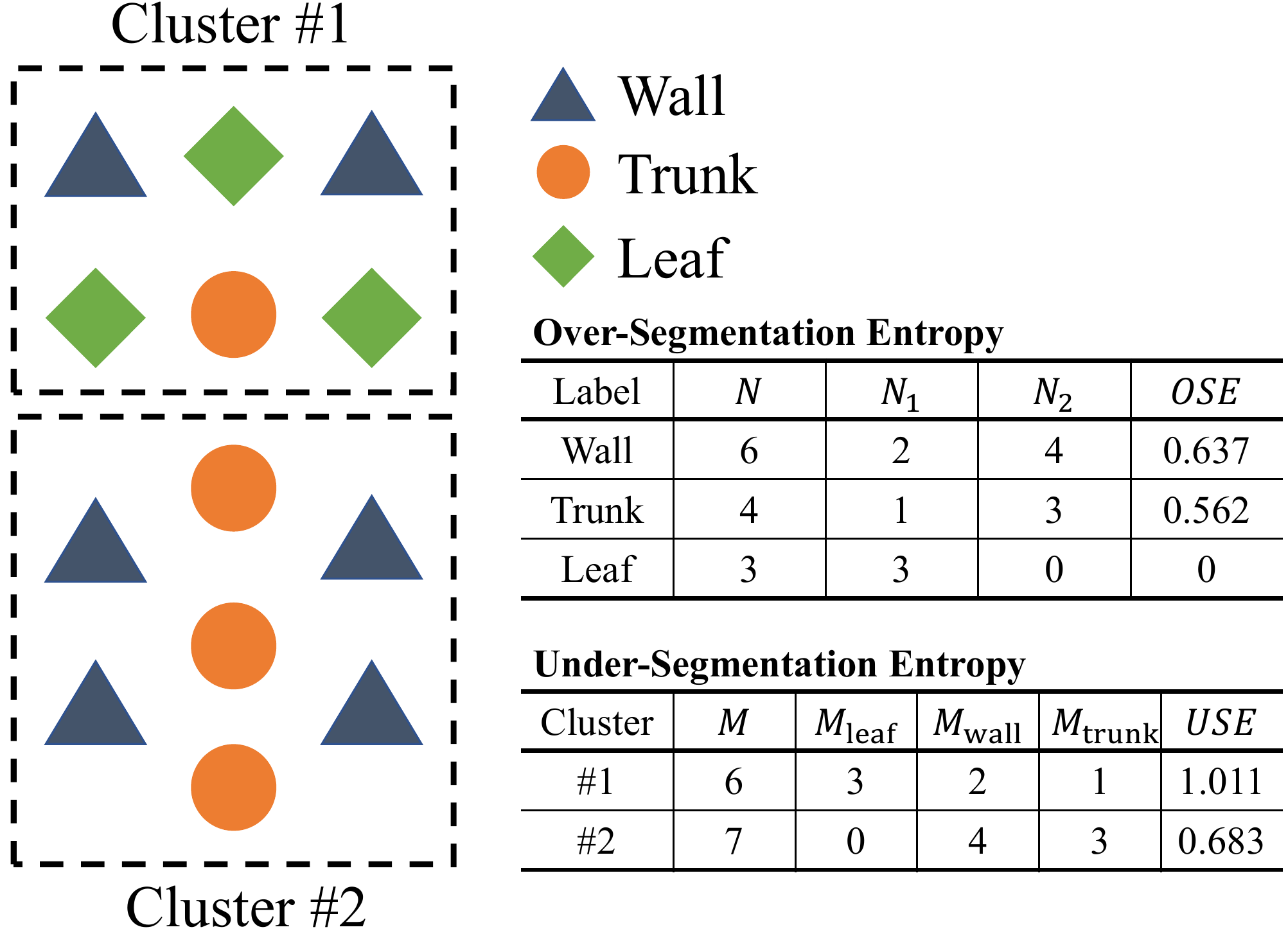}
    \caption{The example illustration of over-segmentation entropy (OSE) and under-segmentation entropy (USE).}
    \label{fig:entropy_example}
    \vspace{-0.5cm}
\end{figure}

Yang \textit{et al.} \cite{node} was the first to adopt quantitative measures to evaluate the segmentation performance on a labeled dataset. However, the authors only compared the ratio of the largest number of clustered points to the total number of points in a labeled object. This criterion is similar to our OSE but does not take into account the effect of multi-instance prediction. In addition, the authors did not suggest under-segmentation evaluation of any kind. Held \textit{et al.}~\cite{held2016probabilistic} and Hu \textit{et al.}~\cite{hu2020learning} suggested over-segmentation and under-segmentation error metrics by counting the number of occurrences for each ground-truth object. However, they fail to capture the distribution of the multi-class labels in the metrics. On the contrary, our two measures are effectively designed so that they can reflect the performance of multi-class prediction.
OSE measures the confusion caused by two or more cluster labels in one ground-truth object, whereas USE measures the confusion caused by two or more ground-truth labels inside one prediction cluster. Suppose $N$ points are measured from an object and, out of $N$ points, $N_i$ points are clustered as label $i$ such that $\sum_{i}(\frac{N_i}{N})=1$. The OSE is calculated by
\begin{equation}
    \text{OSE} =-\sum_{i}\left(\frac{N_i}{N}\right)\log(\frac{N_i}{N}).
\end{equation}
Similarly, suppose $M$ points are clustered as a cluster and, among $M$ points, $M_i$ points have the ground-truth label $i$ such that $\sum_{i}(\frac{M_i}{M})=1$. The USE is calculated by

\begin{equation}
    \text{USE} =-\sum_{i}\left(\frac{M_i}{M}\right)\log(\frac{M_i}{M}).
\end{equation}
Both entropies increase as more conflicting labels exist in an object or in a cluster but reach 0 if only one label exists. Fig.~\ref{fig:entropy_example} illustrates a simple example of OSE and USE. The validity of the two suggested metrics can be seen in Table~\ref{tab:metric_validity}, as the vanilla EC, which is close to the ground-truth, results in lower OSE and USE. However, as the number of points grows, EC is proven to be impractical in computation time~\cite{zermas2017fast}.
\vspace{-0.1cm}
\begin{table}[h!]
    \vspace{-0.1cm}
    \captionsetup{font=footnotesize}
    \centering
    \caption{Average USE and OSE on the sample CARLA dataset}
    \begin{tabular}{l||c|c|c}
    \hline
    Dataset & \multicolumn{3}{c}{Sample CARLA dataset (50 frames)} \\
    \hline
    Metrics & USE $\downarrow$ & OSE $\downarrow$ & Time ($ms$) \\
    \hline 
    Vanilla EC & 0.94 & 4.64 & 7,438 \\
    Proposed & 17.68 & 22.84 & 46 \\
    \hline
    \end{tabular}
    \label{tab:metric_validity}
    \vspace{-0.2cm}
\end{table}

\section{Results and Discussion} \label{sec:result_and_discussion}
\vspace{-0.2cm}
Our proposed algorithm is evaluated in two separate categories: traversable ground segmentation and above-ground object segmentation.
\vspace{-0.2cm}
\subsection{Parameter Studies}
\begin{figure}[t!]
    \captionsetup{font=footnotesize}
    \centering
    \begin{subfigure}[t]{0.32\columnwidth}
        \centering
        \includegraphics[width=\textwidth]{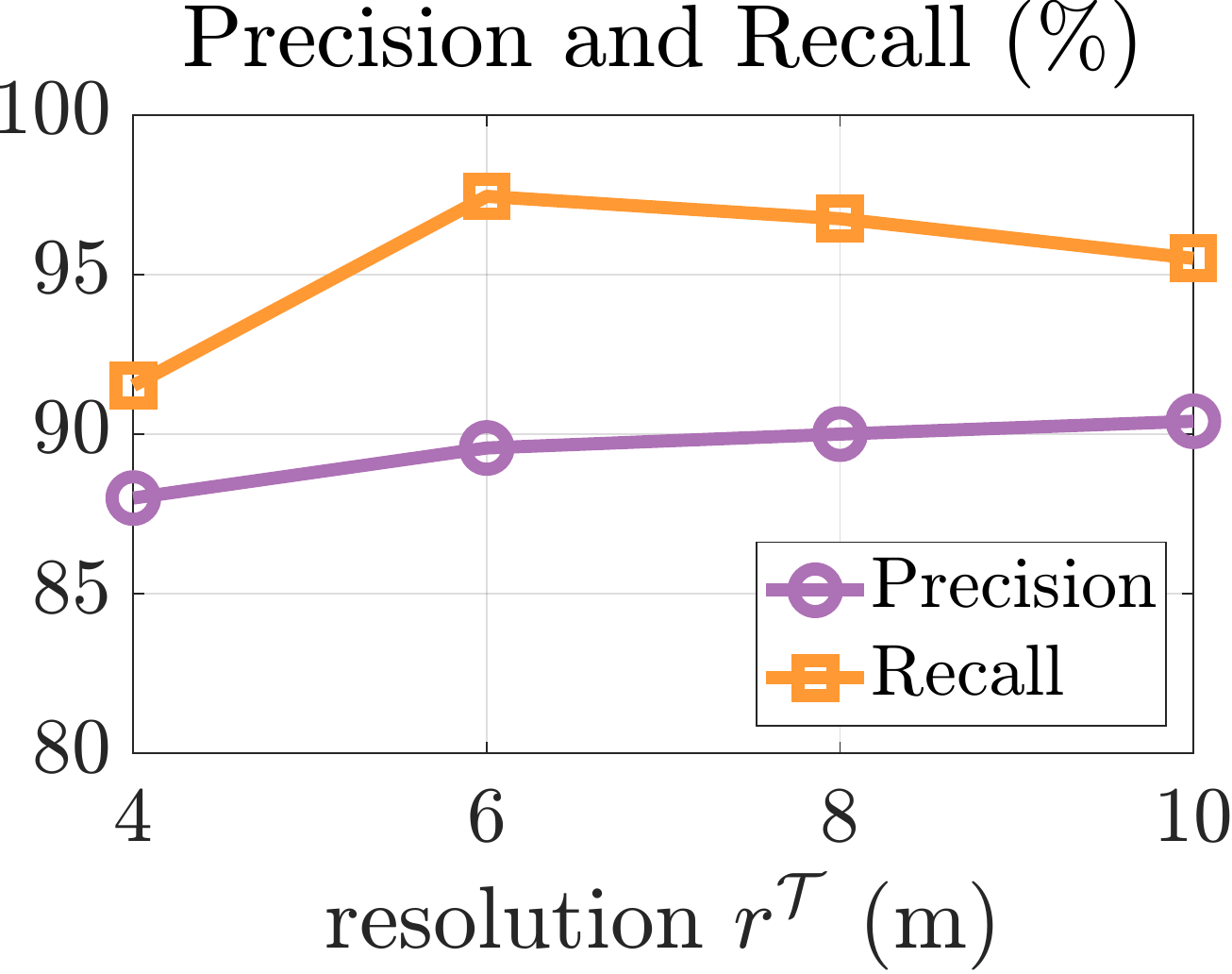}
        \caption{}
    \end{subfigure}
    \begin{subfigure}[t]{0.32\columnwidth}
        \centering
        \includegraphics[width=\textwidth]{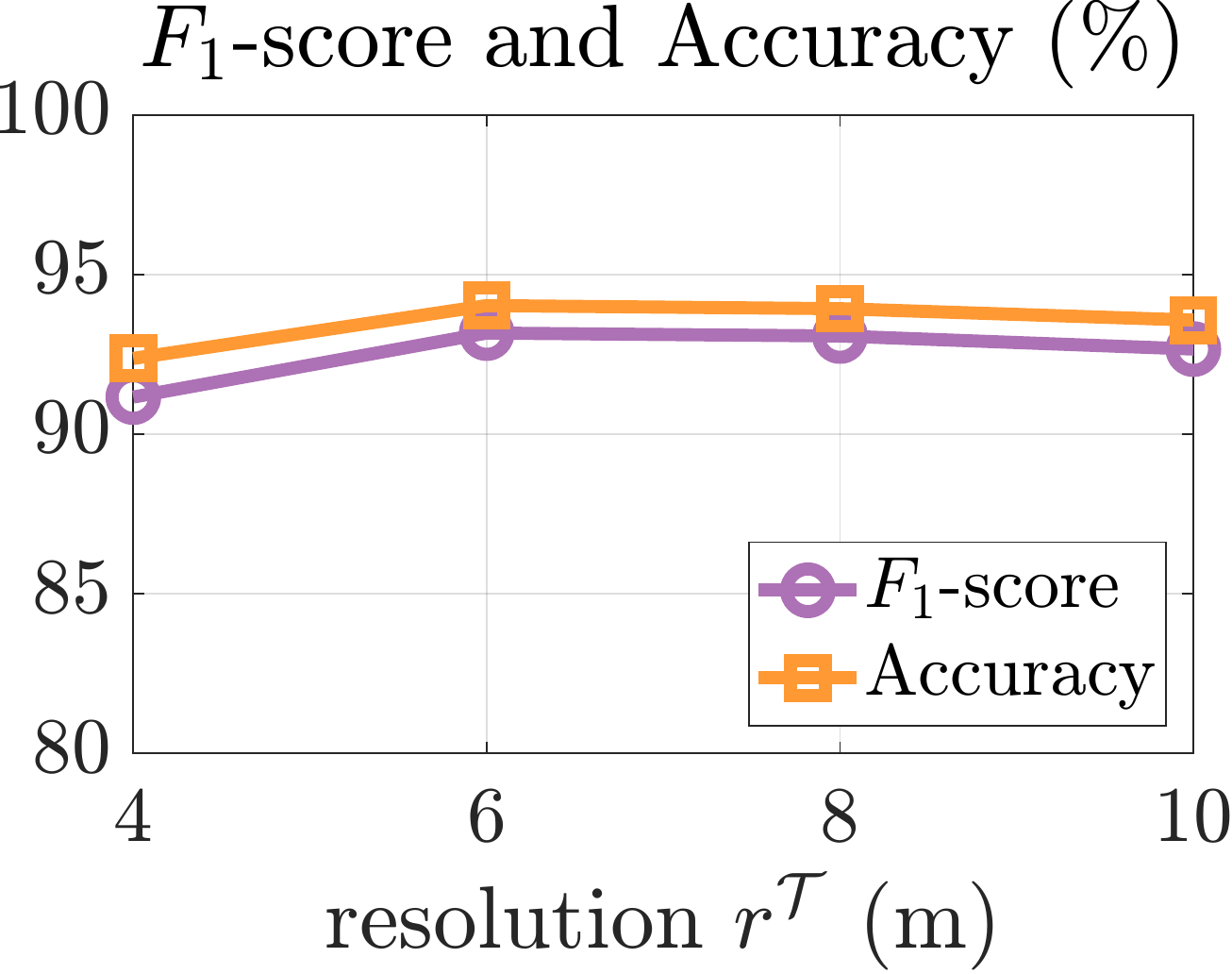}
        \caption{}
    \end{subfigure}
    \begin{subfigure}[t]{0.32\columnwidth}
        \centering
        \includegraphics[width=\textwidth]{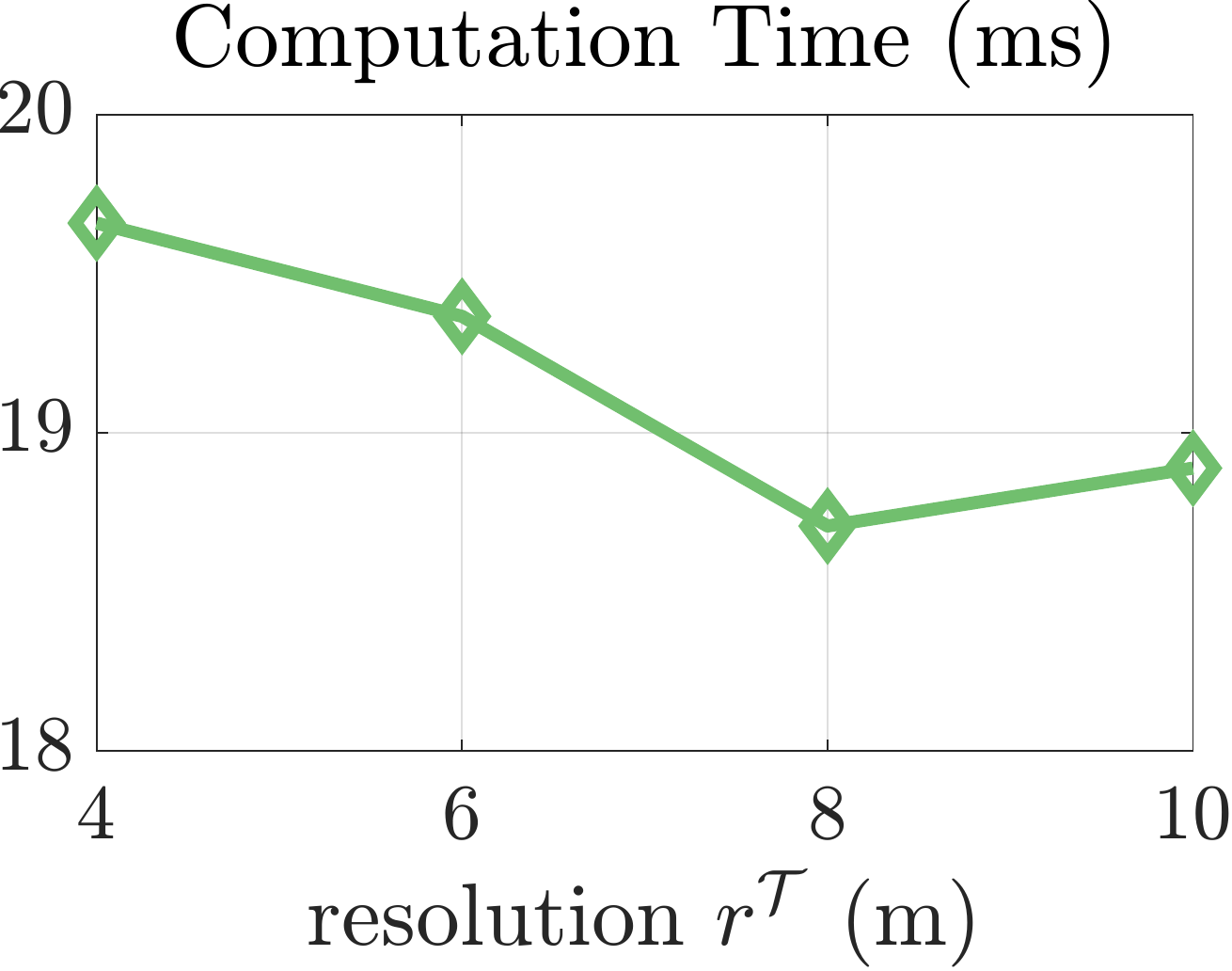}
        \caption{}
    \end{subfigure}
    \vspace{-0.15cm}
    \caption{The effect of $r^{\mathcal{T}}$ on (a) precision, recall, (b) $F_1$-score, accuracy, and (c) computation time.}
    \label{fig:resolution}
    \vspace{-0.4cm}
\end{figure}

\begin{figure}[h!]
    \captionsetup{font=footnotesize}
    \centering
    \begin{subfigure}[t]{0.32\columnwidth}
        \centering
        \includegraphics[width=\textwidth]{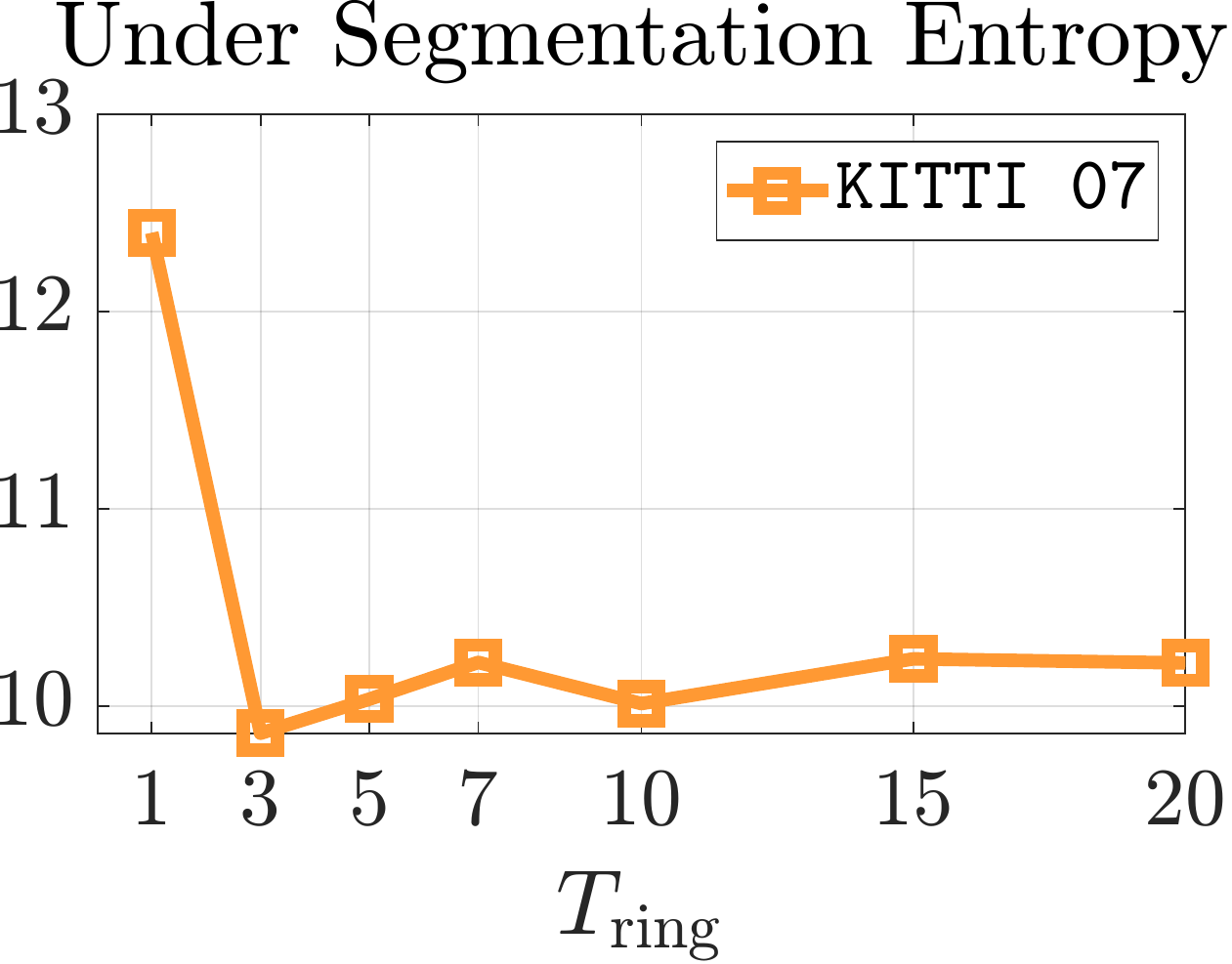}
        \caption{}
        \label{fig:T_ring_USE}
    \end{subfigure}
    \begin{subfigure}[t]{0.32\columnwidth}
        \centering
        \includegraphics[width=\textwidth]{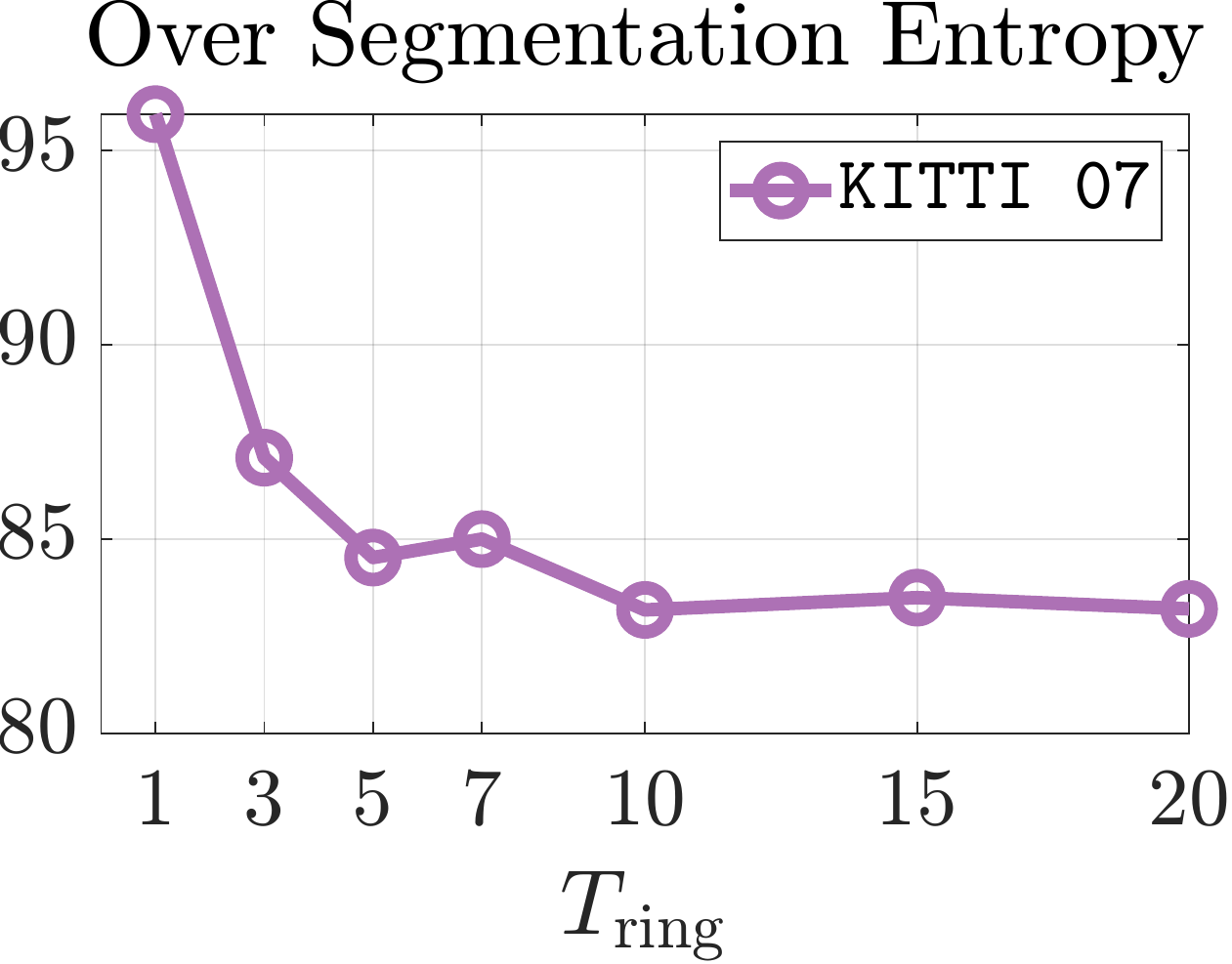}
        \caption{}
        \label{fig:T_ring_OSE}
    \end{subfigure}
    \begin{subfigure}[t]{0.32\columnwidth}
        \centering
        \includegraphics[width=\textwidth]{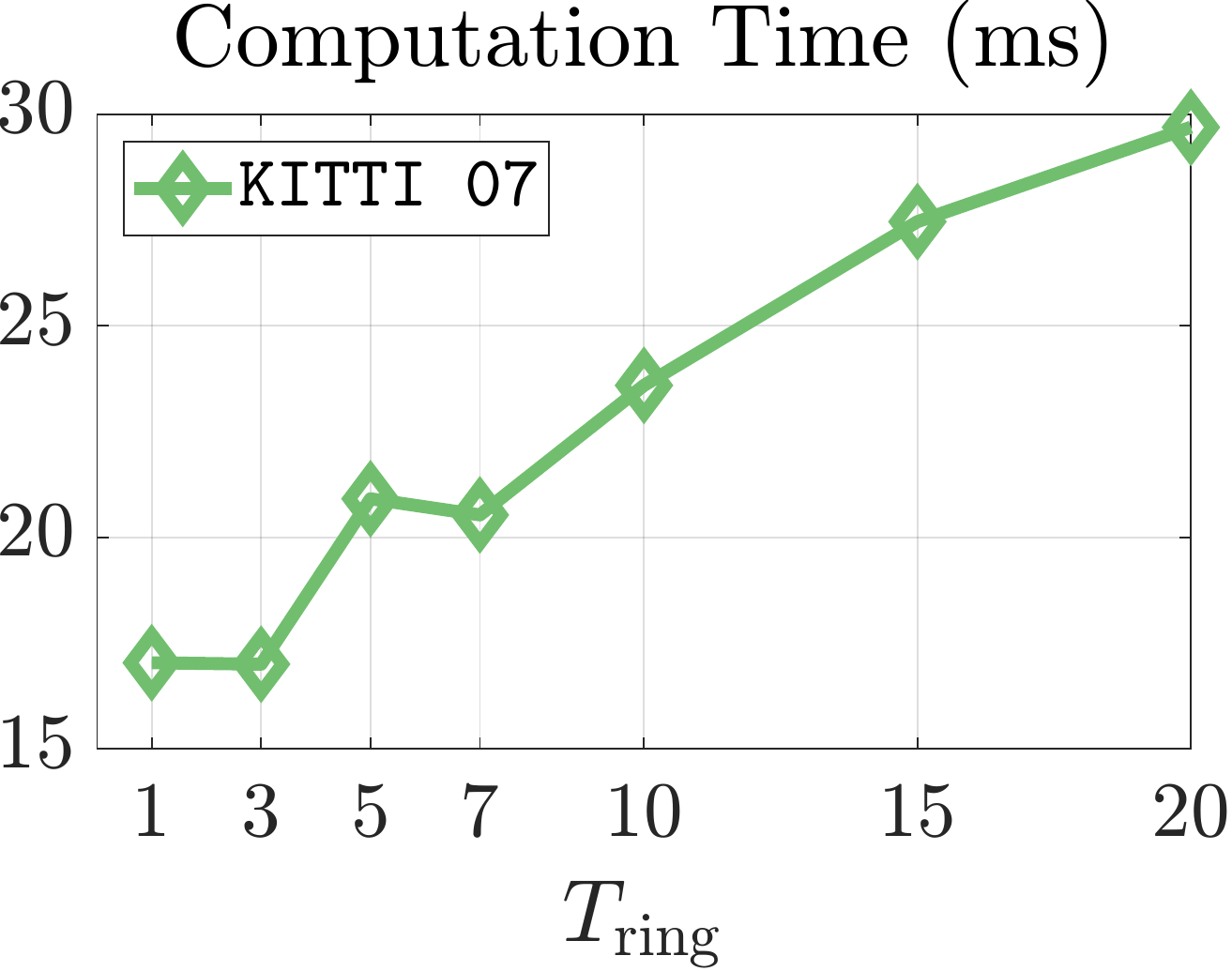}
        \caption{}
        \label{fig:T_ring_times}
    \end{subfigure}
    \vspace{-0.15cm}
    \caption{The effect of $T_{\text{ring}}$ on (a) the under-segmentation entropy, (b) the over-segmentation entropy, and (c) computation time.}
    \label{fig:T_ring_ablation}
    \vspace{-0.4cm}
\end{figure}

\begin{figure}[h!]
    \captionsetup{font=footnotesize}
    \centering
    \begin{subfigure}[t]{0.32\columnwidth}
        \centering
        \includegraphics[width=\textwidth]{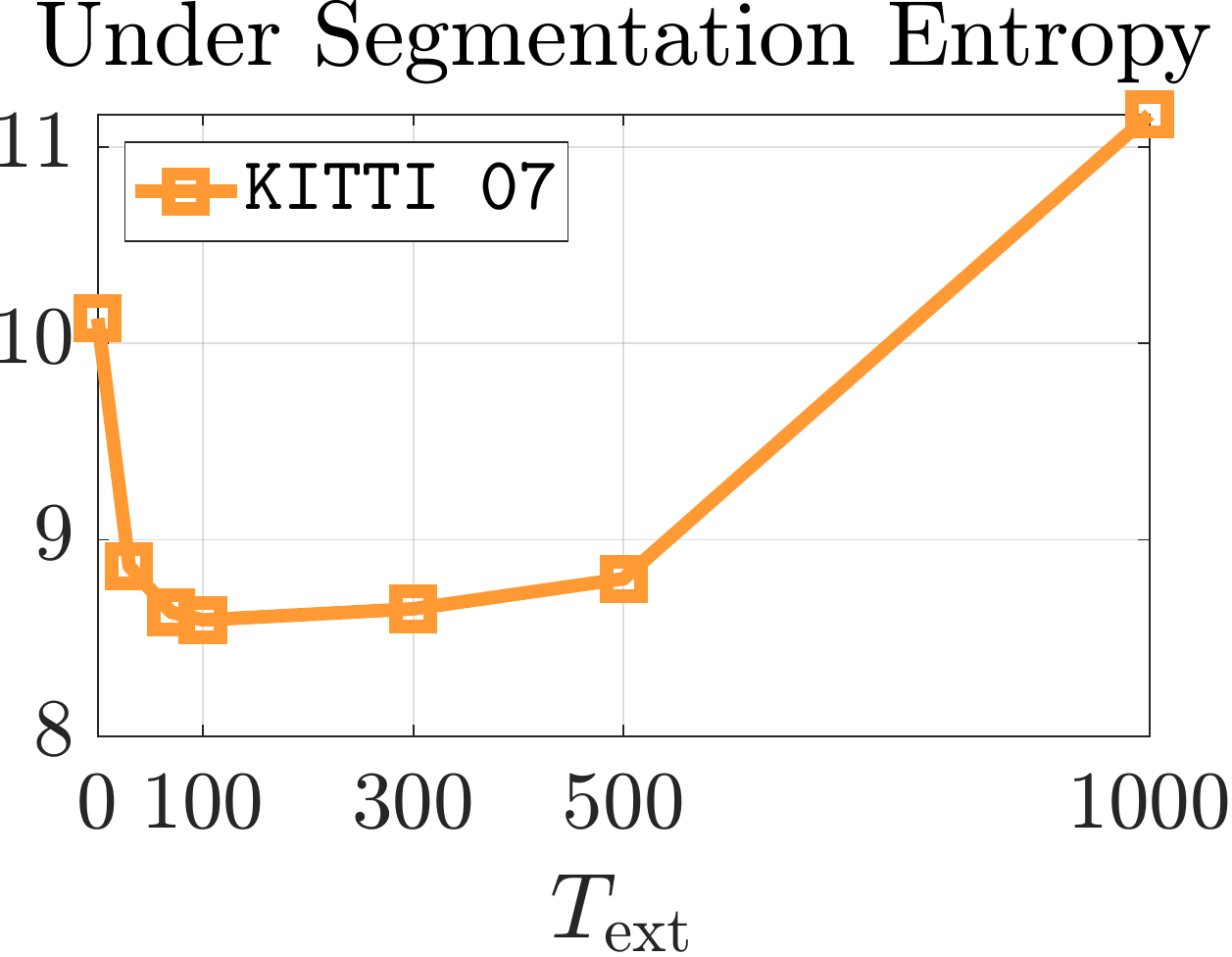}
        \caption{}
        \label{fig:T_ring_USE}
    \end{subfigure}
    \begin{subfigure}[t]{0.32\columnwidth}
        \centering
        \includegraphics[width=\textwidth]{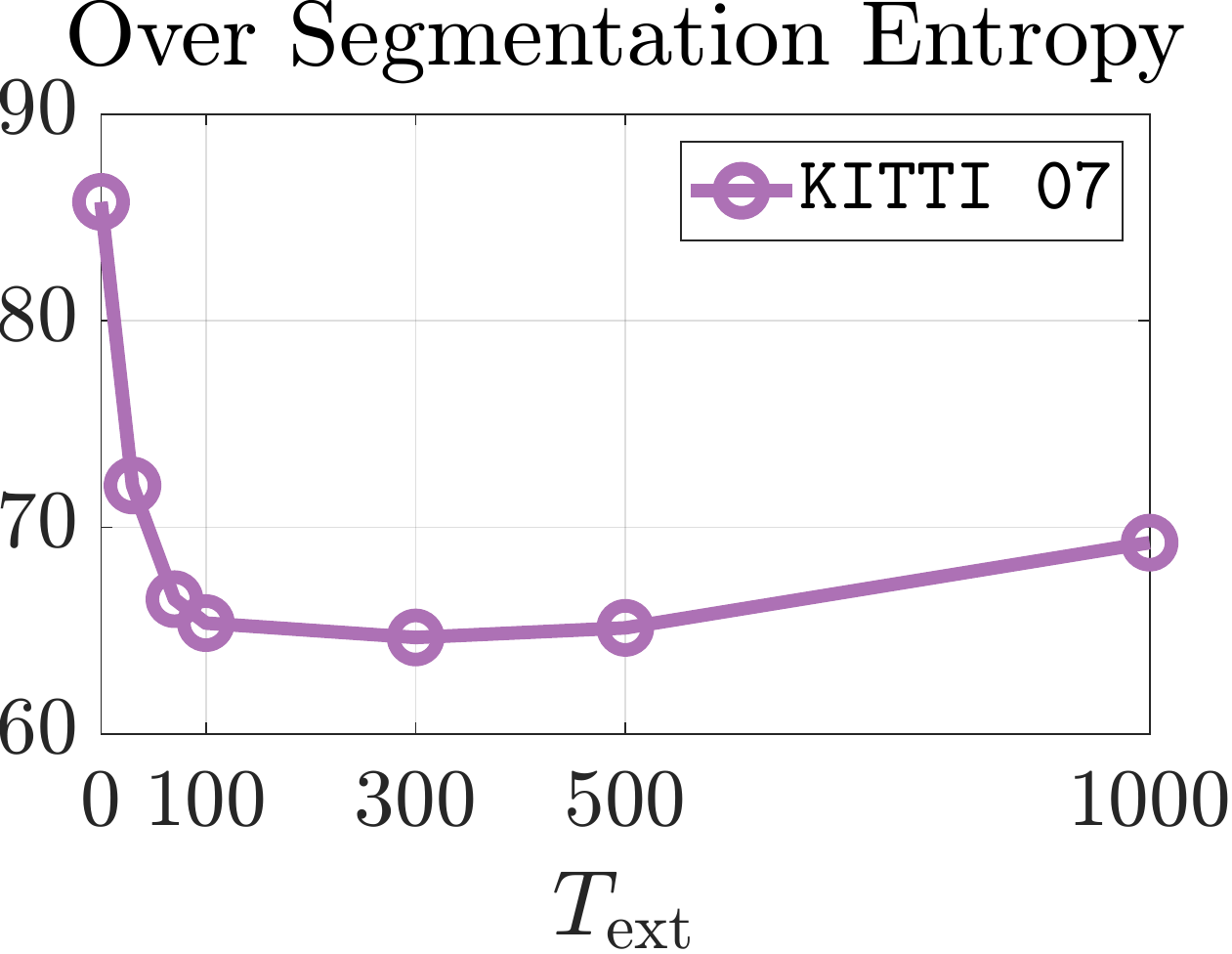}
        \caption{}
        \label{fig:T_ring_OSE}
    \end{subfigure}
    \begin{subfigure}[t]{0.32\columnwidth}
        \centering
        \includegraphics[width=\textwidth]{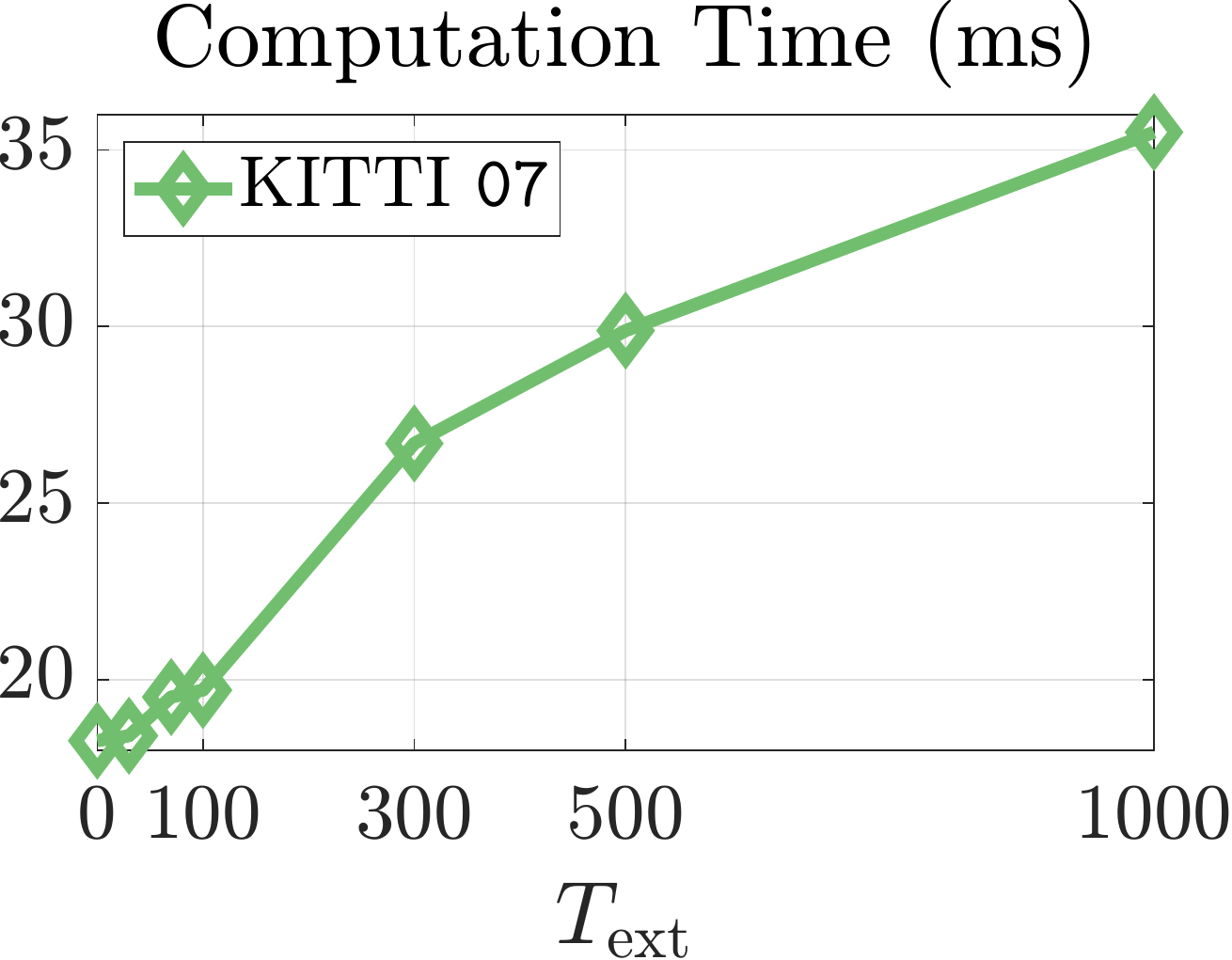}
        \caption{}
        \label{fig:T_ring_times}
    \end{subfigure}
    \vspace{-0.15cm}
    \caption{The effect of $T_{\text{ext}}$ on (a) the under-segmentation entropy, (b) the over-segmentation entropy, and (c) computation time.}
    \label{fig:T_ext_ablation}
    \vspace{-0.7cm}
\end{figure}

We first shed light on the effect of some important parameters on the performance using sequence \texttt{07} of Semantic KITTI dataset. This dataset is carefully chosen for the studies since it has the most varying urban conditions so that we can generalize the parameters. Figs.~\ref{fig:resolution}~-~\ref{fig:T_ext_ablation} show the effect of tri-grid resolution $r^{\mathcal{T}}$, \textcolor{rv}{ring-wise search window size} $T_\text{ring}$, and ${T_\text{ext}}$, respectively. The other parameters are kept constant as in Table~\ref{tab:parems} during the experiment of each parameter.    

\subsubsection{The effect of $r_{\mathcal{T}}$}

With a small resolution, the performance of ours worsens. This is because the number of points included in each node is insufficient for planar modeling. On the other hands, there is no significant change in the performance at a resolution of 6 m or higher. Considering the computation time as well, we found that a resolution of 8 meter would suffice.

\subsubsection{The effect of $T_{\text{ring}}$}
A large $T_{\text{ring}}$ allows a larger search space for the vertical linkages between nodes. Since the computation time increases with an increasing $T_{\text{ring}}$, we have to carefully determine the optimal value. From Fig. \ref{fig:T_ring_ablation}, we figured that the values within 3 to 5 are reasonable for $T_{\text{ring}}$.

\subsubsection{The effect of $T_{\text{ext}}$}
$T_{\text{ext}}$ helps the algorithm extract more overlapping node candidates to merge vertically. The larger $T_{\text{ext}}$ is, the less likely the vertical separation will occur. As shown in Fig.~\ref{fig:T_ext_ablation}, thanks to the binary search explained in Section~\ref{ssec:vert_update}, the increased $T_{\text{ext}}$ does not significantly increase the computation burden. However, the performance of the segmentation saturates or worsens after 100. As a result, we chose 100 as an optimal value for $T_{\text{ext}}$.

\subsection{Traversable Ground Segmentation}
\begin{figure}[t!]
    \captionsetup{font=footnotesize}
    \centering
    \begin{subfigure}[t]{\columnwidth}
        \centering
        \includegraphics[width=0.45\textwidth]{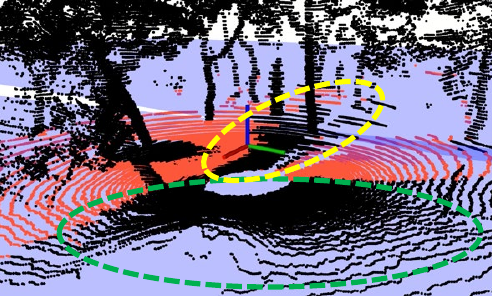}
        \includegraphics[width=0.45\textwidth]{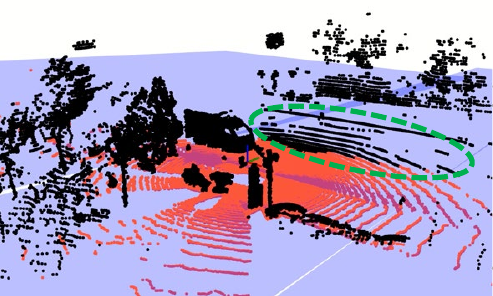}
        \caption{Zermas \textit{et al.}~\cite{zermas2017fast}}
    \end{subfigure}
    \begin{subfigure}[t]{\columnwidth}
        \centering
        \includegraphics[width=0.45\textwidth]{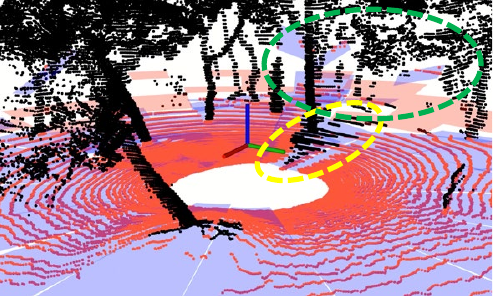}
        \includegraphics[width=0.45\textwidth]{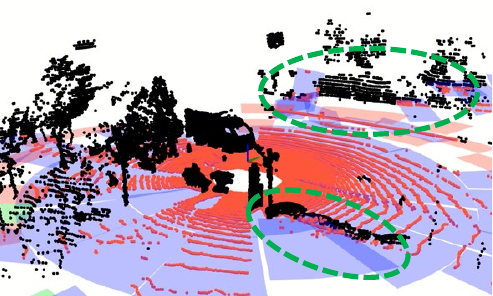}
        \caption{Lim \textit{et al.}~\cite{lim21patchwork}}
    \end{subfigure}
    \begin{subfigure}[t]{\columnwidth}
        \centering
        \includegraphics[width=0.45\textwidth]{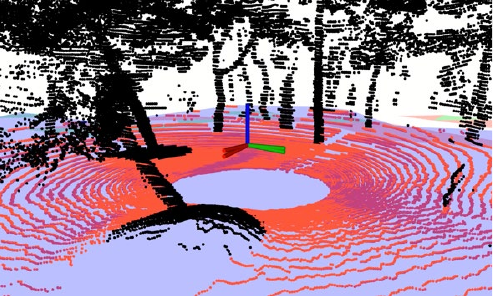}
        \includegraphics[width=0.45\textwidth]{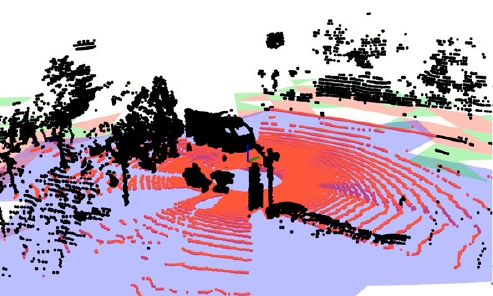}
        \caption{Proposed}
    \end{subfigure}
    \vspace{-0.1cm}
    \caption{Ground segmentation results on bumpy and flat but tilted terrains in the rough terrain dataset. The estimated ground and object points are colored in red and black, respectively. The yellow dashed ellipses indicate the segmentation failures due to skewed scan data, and the green ones show the failures in plane modeling of the ground. The estimated ground points by \cite{zermas2017fast} are fallen into a local minimum, and the discontinuous planes, modeled by \cite{lim21patchwork}, cause the local failures, such as detecting a wall as the ground.}
    \label{fig:rough_terrain_gseg}
    \vspace{-0.6cm}
\end{figure}

\vspace{-0.1cm}
Table~\ref{tab:quantitative_comparison_gseg} shows that our proposed algorithm demonstrates the highest $F_{1}$-score and accuracy with the lowest perturbation and computation time in both datasets. Also, to further examine the robustness of the proposed algorithm, three other algorithms, which show high performance on both datasets, are compared on the bumpy terrain with a myriad of low-lying obstacles and trees as shown in Fig.~\ref{fig:rough_terrain_gseg}. 

Our ground segmentation considering the traversability allows not only to model the bumpy terrains but also to extract the low-lying obstacles such as low stairs and vegetations. On the other hand, since Ground Plane Fitting (GPF)~\cite{zermas2017fast} divides areas only in the moving direction, the estimated ground tends to converge to a local minimum on the uneven terrain. Patchwork~\cite{lim21patchwork} shows relatively decent segmentation results on ground modeling by taking into account the characteristics of LiDAR data distribution. Nevertheless, its lack of interaction between the bins causes the discontinuity of the planar models, thus making it difficult to extract the low-lying obstacles from the ground. As a result, our design of TGF, which focuses on the geometric relationship between terrain elements via a graph structure, assists in the accurate modeling of varying-sloped planes and makes our algorithm robust to both urban and wild environments.

\begin{table}[h!]
    \vspace{-0.1cm}
    \captionsetup{font=footnotesize}
    \centering
    \caption{Quantitative comparison for the traversable ground segmentation on CARLA simulator and whole sequences of the Semantic KITTI dataset. Units are $ms$ for T (computation time) and \%, otherwise. $\mu$ and $\sigma$ are the mean and stdev. of each metric, respectively.}
    \vspace{-0.15cm}
    \setlength{\tabcolsep}{4.2pt}
    \begin{tabular}{l||c|c|c|c|c|c|c}
        \hline
        \multirow{2}{*}{Metrics}& P & R & \multicolumn{2}{c|}{$F_{1}$-score} & \multicolumn{2}{c|}{Accuracy}& T\\
        & $\mu$ $\uparrow$ & $\mu$ $\uparrow$ & $\mu$ $\uparrow$ & $\sigma$ $\downarrow$ & $\mu$ $\uparrow$ & $\sigma$ $\downarrow$ & $\mu$ \\
        \hline
        \hline
        Dataset& \multicolumn{7}{c}{CARLA}\\
        \hline
        RANSAC \cite{fischler1981ransac}                    & 88.3          & 92.1 & 89.0   & 20.0          &  88.5 & 21.9 & 67\\
        Zermas \textit{et al.}~\cite{zermas2017fast}        & 97.5          & 92.7 & 94.3   & 10.3          &  95.0 & ~7.9 & 20\\  
        Narksri \textit{et al.}~\cite{narksri2018slope}     & \textbf{97.7} & 83.3 & 89.6   & ~8.2          &  60.7 & ~8.8 & 19\\ 
        Lim \textit{et al.}~\cite{lim21patchwork}           & 94.2          & 95.6 & 94.8   & ~3.9          &  93.6 & ~3.8 & 28\\
        Proposed w/o TTMF                                   & 95.8          & 97.0 & 96.3   & ~2.8          & 93.4 & ~2.9 & 14 \\
        Proposed                                            & 96.6          & \textbf{97.2} & \textbf{96.7} & \textbf{~2.7} &  \textbf{95.9} & \textbf{~2.5} & \textbf{14}\\
        \hline
        \hline
        Dataset& \multicolumn{7}{c}{Semantic KITTI}\\
        \hline
        RANSAC \cite{fischler1981ransac}                    & 82.7 & 94.0 & 87.2 & 15.4 & 88.4 & 13.0 &  83  \\
        Zermas \textit{et al.}~\cite{zermas2017fast}        & \textbf{91.4} & 83.9 & 85.6 & 18.3 & 88.9 & 12.3 &  23  \\
        Narksri \textit{et al.}~\cite{narksri2018slope}     & 89.1 & 74.0 & 80.6 & 10.7 & 81.4 & ~6.8 &  75  \\ 
        Lim \textit{et al.}~\cite{lim21patchwork}           & 87.5 & \textbf{97.6} & 92.1 & ~4.5 & 92.0 & ~4.4 &  26  \\
        Proposed w/o TTMF                                   & 88.6 & 96.4 & 92.2 & ~4.5 & 93.3 & ~3.5 & 18 \\
        Proposed                                            & 90.0 & 96.7 & \textbf{93.1} & \textbf{~4.3} & \textbf{93.9} & \textbf{~3.7} &  \textbf{19}  \\
        \hline
    \end{tabular}
    \label{tab:quantitative_comparison_gseg}
    \vspace{-0.5cm}
\end{table}

\vspace{-0.1cm}
\subsection{Above-Ground Object Segmentation}
\begin{figure*}[t!]
    \captionsetup{font=footnotesize}
    \centering
    \begin{subfigure}[t]{0.97\textwidth}
        \centering
        \includegraphics[width=\textwidth]{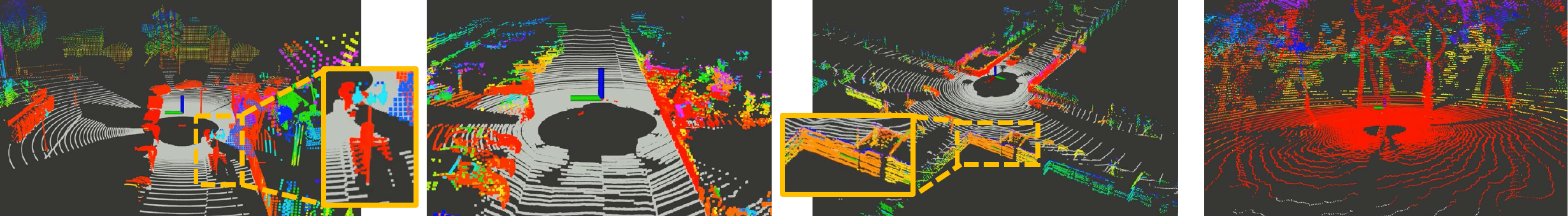}
        \caption{GPF + SLR~\cite{zermas2017fast}}
        \label{fig:result_dummy_a}
    \end{subfigure}
    \begin{subfigure}[t]{0.97\textwidth}
        \centering
        \includegraphics[width=\textwidth]{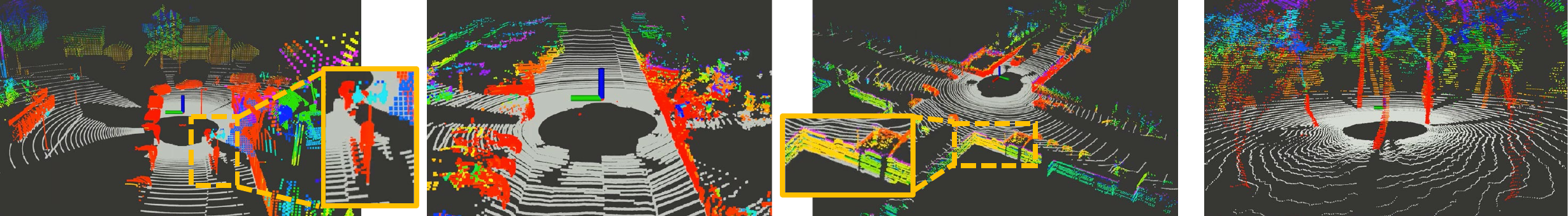}
        \caption{TGS + SLR}
        \label{fig:result_dummy_b}
    \end{subfigure}
    \begin{subfigure}[t]{0.97\textwidth}
        \centering
        \includegraphics[width=\textwidth]{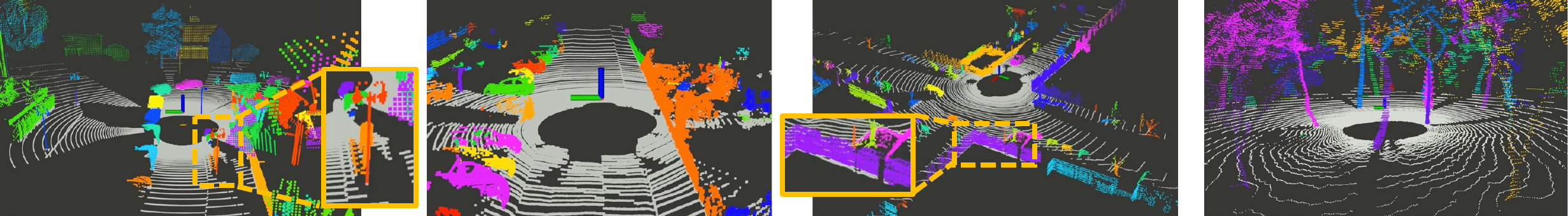}
        \caption{Proposed}
        \label{fig:result_dummy_c}
    \end{subfigure}
    \vspace{-0.15cm}
    \caption{Qualitative comparisons on CARLA simulator, sequence \texttt{05} and \texttt{07} of Semantic KITTI datasets, and rough terrain dataset in the order from left to right column. The estimated ground points are shown in grey, and the segmented objects are represented by non-grey colors. TRAVEL overcomes the vertical separation problem on the pole as in the first column and the horizontal separation problem of the wall as in the third column.}
    \label{fig:result_dummy}
    \vspace{-0.7cm}
\end{figure*}

\vspace{-0.1cm}
Table~\ref{tab:quantitative_comparison_oseg} shows that our proposed method demonstrates the lowest USE and OSE in both datasets. As illustrated in Fig.~\ref{fig:result_dummy}, the pole-like object, which is prone to vertical separation, is well clustered without over-segmentation by our inter-ring linkages. Also, the horizontal separations of the walls by occlusion are largely prevented by our skipping linkages. Note that TGS (ours) + Scan Line Run (SLR)~\cite{zermas2017fast} increases the performance significantly compared with GPF + SLR. This performance boost implies that our traversable ground segmentation outperforms GPF by a large margin.

Naturally, SLR should perform faster than our algorithm, since SLR requires $\mathit{O}(n)$ while ours requires $\mathit{O}(\mathit{N}\log(\mathit{N}))$ during the vertical update, where $n$ is the number of points and $N$ is the number of nodes. Therefore, as expected, SLR runs faster than ours in the Semantic KITTI dataset. Nevertheless, as more scan points can be compactly represented as a node, our algorithm can run faster in some cases. For instance, in the CARLA simulation dataset, where scan points are measured with less noise and thus can be compactly represented, $N$ might have become small enough to beat $\mathit{O}(n)$ complexity of the SLR. Moreover, the naive indexing of nodes in two different rings in SLR, which requires $\mathit{O}(n)$, induces serious vertical separation in a noisy environment. In conclusion, our algorithm notably outperforms the state-of-the-art algorithm at the cost of time complexity while still maintaining real-time performance by about 20-30 Hz.  

\begin{table}[t!]
    \vspace{-0.1cm}
    \captionsetup{font=footnotesize}
    \centering
    \caption{Quantitative comparison for the above-ground object segmentation on CARLA simulator and whole sequences of the Semantic KITTI dataset.}
    \begin{tabular}{l||c|c|c|c|c}
        \hline
        \multirow{2}{*}{Metrics}& \multicolumn{2}{c|}{USE} & \multicolumn{2}{c|}{OSE} & T\\
        & $\mu$ $\downarrow$ & $\sigma$ $\downarrow$ & $\mu$ $\downarrow$ & $\sigma$ $\downarrow$ & $\mu$ $\downarrow$\\
        \hline
        \hline
        Dataset& \multicolumn{5}{c}{CARLA}\\
        \hline
        GPF + SLR \cite{zermas2017fast}       
            & 67.01 & 64.31 & 315.01 & 114.61 &  36  \\
        TGS + SLR                                   
            & 37.59 & 28.28 & 230.48 & ~83.63 &  35  \\
        Proposed                                  
            & \textbf{~7.86} & \textbf{~4.99} & \textbf{~31.71} & \textbf{~15.88} &  \textbf{32}  \\
        \hline
        \hline
        Dataset& \multicolumn{5}{c}{Semantic KITTI}\\
        \hline
        GPF + SLR \cite{zermas2017fast}       
            & 117.08 & 127.40 & 301.16 & 114.29 &  43  \\
        TGS + SLR                                   
            & ~45.33 & ~22.14 & 228.53 & ~86.13 &  \textbf{37}  \\
        Proposed                                  
            & \textbf{~24.07} & \textbf{~11.88} & \textbf{~70.40} & \textbf{~34.44} &  50  \\
        \hline
    \end{tabular}
    \label{tab:quantitative_comparison_oseg}
    \vspace{-0.5cm}
\end{table}

\section{Conclusions} \label{sec:conclusion}
\vspace{-0.2cm}
In this study, we propose a two-step segmentation of traversable ground and above-ground objects, TRAVEL. Our node-edge representation of a point cloud allows accurate modeling of the ground and efficient searching of neighboring points. Our traversable ground segmentation outperforms the prior studies in terms of conventional evaluation metrics. Also, our object segmentation brings about less under-segmentation and over-segmentation, which are assessed using our newly proposed metrics. In essence, unlike learning-based methods, our algorithm cannot assign the same class label, for instance, to two spatially distant walls separated by a large object; $T_{\text{skip}} $ mitigates separation of the walls caused by a thin object, yet $T_{\text{horz}}$ may limit the merging of them. Due to this limitation, our work is concerned more with real-time navigation by spotting relevant targets under unseen environments, than with classifying targets with specific class labels. As future works, we would like to apply the proposed algorithm to the navigation task, tracking the relevant objects to identify and remove dynamic motion from a scene.

\vspace{-0.1cm}
\bibliographystyle{IEEEtran}
\vspace{-0.5cm}
\bibliography{./iros22}

\begin{thebibliography}{10}
\providecommand{\url}[1]{#1}
\csname url@rmstyle\endcsname
\providecommand{\newblock}{\relax}
\providecommand{\bibinfo}[2]{#2}
\providecommand\BIBentrySTDinterwordspacing{\spaceskip=0pt\relax}
\providecommand\BIBentryALTinterwordstretchfactor{4}
\providecommand\BIBentryALTinterwordspacing{\spaceskip=\fontdimen2\font plus
\BIBentryALTinterwordstretchfactor\fontdimen3\font minus
  \fontdimen4\font\relax}
\providecommand\BIBforeignlanguage[2]{{%
\expandafter\ifx\csname l@#1\endcsname\relax
\typeout{** WARNING: IEEEtran.bst: No hyphenation pattern has been}%
\typeout{** loaded for the language `#1'. Using the pattern for}%
\typeout{** the default language instead.}%
\else
\language=\csname l@#1\endcsname
\fi
#2}}

\bibitem{pan2021mulls}
Y.~Pan, P.~Xiao, Y.~He, Z.~Shao, and Z.~Li, ``{MULLS: Versatile LiDAR SLAM via
  multi-metric linear least square},'' in \emph{Proc. IEEE Int. Conf. on Robot.
  and Automat.}, 2021, pp. 11\,633--11\,640.

\bibitem{sung2021if}
C.~Sung, S.~Jeon, H.~Lim, and H.~Myung, ``{What if there was no revisit?
  Large-scale graph-based SLAM with traffic sign detection in an HD map using
  LiDAR inertial odometry},'' \emph{Intell. Service Robot.}, pp. 1--10, 2021.

\bibitem{arora2021mapping}
M.~Arora, L.~Wiesmann, X.~Chen, and C.~Stachniss, ``{Mapping the static parts
  of dynamic scenes from 3D LiDAR point clouds exploiting ground
  segmentation},'' in \emph{Proc. European Conf. on Mobile Robots (ECMR)},
  2021, pp. 1--6.

\bibitem{yoon2019mapless}
D.~Yoon, T.~Tang, and T.~Barfoot, ``{Mapless online detection of dynamic
  objects in 3D LiDAR},'' in \emph{Proc. Conf. on Comput. and Robot Vis,
  (CRV)}, 2019, pp. 113--120.

\bibitem{dosovitskiy17}
A.~Dosovitskiy, G.~Ros, F.~Codevilla, A.~Lopez, and V.~Koltun, ``{CARLA}: {An}
  open urban driving simulator,'' in \emph{Proc. Conf. on Robot Learning},
  2017, pp. 1--16.

\bibitem{behley2019semantickitti}
J.~Behley, M.~Garbade, A.~Milioto, J.~Quenzel, S.~Behnke, C.~Stachniss, and
  J.~Gall, ``{SemanticKITTI: A dataset for semantic scene understanding of
  LiDAR sequences},'' in \emph{Proc. IEEE/CVF Int. Conf. Comput. Vis. (ICCV)},
  2019, pp. 9297--9307.

\bibitem{fischler1981ransac}
M.~A. Fischler and R.~C. Bolles, ``Random sample consensus: A paradigm for
  model fitting with applications to image analysis and automated
  cartography,'' \emph{Commun. ACM}, vol.~24, no.~6, pp. 381--395, 1981.

\bibitem{narksri2018slope}
P.~Narksri, E.~Takeuchi, Y.~Ninomiya, Y.~Morales, N.~Akai, and N.~Kawaguchi,
  ``{A slope-robust cascaded ground segmentation in 3D point cloud for
  autonomous vehicles},'' in \emph{Proc. IEEE Int. Conf. on Intell. Transport.
  Syst. (ITSC)}, 2018, pp. 497--504.

\bibitem{zermas2017fast}
D.~Zermas, I.~Izzat, and N.~Papanikolopoulos, ``{Fast segmentation of 3D point
  clouds: A paradigm on LiDAR data for autonomous vehicle applications},'' in
  \emph{Proc. IEEE Int. Conf. Robot. Automat.}, 2017, pp. 5067--5073.

\bibitem{lim21patchwork}
H.~Lim, M.~Oh, and H.~Myung, ``{Patchwork: Concentric zone-based region-wise
  ground segmentation with ground likelihood estimation using a 3D LiDAR
  sensor},'' \emph{IEEE Robot. Automat. Lett.}, vol.~6, no.~4, pp. 6458--6465,
  2021.

\bibitem{xue2021lidar}
H.~Xue, H.~Fu, R.~Ren, J.~Zhang, B.~Liu, Y.~Fan, and B.~Dai, ``{LiDAR-based
  drivable region detection for autonomous driving},'' in \emph{Proc. IEEE/RSJ
  Int. Conf. on Intell. Robots and Syst.}, 2021, pp. 1110--1116.

\bibitem{moosmann2009segmentation}
F.~Moosmann, O.~Pink, and C.~Stiller, ``{Segmentation of 3D LiDAR data in
  non-flat urban environments using a local convexity criterion},'' in
  \emph{Proc. IEEE Intell. Veh. Symp. (IVS)}, 2009, pp. 215--220.

\bibitem{bogo}
I.~Bogoslavskyi and C.~Stachniss, ``{Fast range image-based segmentation of
  sparse 3D laser scans for online operation},'' in \emph{Proc. IEEE/RSJ Int.
  Conf. on Intell. Robots and Syst.}, 2016, pp. 163--169.

\bibitem{mesh}
P.~Burger, B.~Naujoks, and H.~Wuensche, ``Fast dual decomposition based
  mesh-graph clustering for point clouds,'' in \emph{Proc. Int. Conf. on
  Intell. Transport. Syst. (ITSC)}, 2018, pp. 1129--1135.

\bibitem{node}
H.~Yang, Z.~Wang, L.~Lin, H.~Liang, W.~Huang, and F.~Xu, ``{Two-layer-graph
  clustering for real-time 3D LiDAR point cloud segmentation},'' \emph{Appl.
  Sci.}, vol.~10, no.~23, p. 8534, 2020.

\bibitem{zhang2020instance}
F.~Zhang, C.~Guan, J.~Fang, S.~Bai, R.~Yang, P.~H. Torr, and V.~Prisacariu,
  ``Instance segmentation of lidar point clouds,'' in \emph{Proc. IEEE Int.
  Conf. on Robot. and Automat.}, 2020, pp. 9448--9455.

\bibitem{wong2020identifying}
K.~Wong, S.~Wang, M.~Ren, M.~Liang, and R.~Urtasun, ``Identifying unknown
  instances for autonomous driving,'' in \emph{Conference on Robot
  Learning}.\hskip 1em plus 0.5em minus 0.4em\relax PMLR, 2020, pp. 384--393.

\bibitem{milioto2019rangenet++}
A.~Milioto, I.~Vizzo, J.~Behley, and C.~Stachniss, ``{RangeNet++: Fast and
  accurate LiDAR semantic segmentation},'' in \emph{Proc. IEEE/RSJ Int. Conf.
  Intell. Robots Syst.}, 2019, pp. 4213--4220.

\bibitem{xu2021rpvnet}
J.~Xu, R.~Zhang, J.~Dou, Y.~Zhu, J.~Sun, and S.~Pu, ``Rpvnet: A deep and
  efficient range-point-voxel fusion network for lidar point cloud
  segmentation,'' in \emph{Proc. IEEE/CVF Int. Conf. on Comput. Vis.}, 2021,
  pp. 16\,024--16\,033.

\bibitem{peng2022mass}
K.~Peng, J.~Fei, K.~Yang, A.~Roitberg, J.~Zhang, F.~Bieder, P.~Heidenreich,
  C.~Stiller, and R.~Stiefelhagen, ``Mass: Multi-attentional semantic
  segmentation of lidar data for dense top-view understanding,'' \emph{Proc.
  IEEE Transac. on Intell. Transport. Syst.}, 2022.

\bibitem{paigwar2020gndnet}
A.~Paigwar, {\"O}.~Erkent, D.~Sierra-Gonzalez, and C.~Laugier, ``Gndnet: Fast
  ground plane estimation and point cloud segmentation for autonomous
  vehicles,'' in \emph{Proc. IEEE/RSJ Int. Conf. on Intell. Robots and Syst.},
  2020, pp. 2150--2156.

\bibitem{shan2020lio}
T.~Shan, B.~Englot, D.~Meyers, W.~Wang, C.~Ratti, and D.~Rus, ``{LIO-SAM:
  Tightly-coupled LiDAR inertial odometry via smoothing and mapping},'' in
  \emph{Proc. IEEE/RSJ Int. Conf. on Intell. Robots and Syst.}, 2020, pp.
  5135--5142.

\bibitem{weinmann2015semantic}
M.~Weinmann, B.~Jutzi, S.~Hinz, and C.~Mallet, ``Semantic point cloud
  interpretation based on optimal neighborhoods, relevant features and
  efficient classifiers,'' \emph{ISPRS Journal of Photo. and Remote Sens.},
  vol. 105, pp. 286--304, 2015.

\bibitem{held2016probabilistic}
D.~Held, D.~Guillory, B.~Rebsamen, S.~Thrun, and S.~Savarese, ``A probabilistic
  framework for real-time 3d segmentation using spatial, temporal, and semantic
  cues.'' in \emph{Robotics: Science and Systems}, vol.~12, 2016.

\bibitem{hu2020learning}
P.~Hu, D.~Held, and D.~Ramanan, ``Learning to optimally segment point clouds,''
  \emph{IEEE Robot. Automat. Lett.}, vol.~5, no.~2, pp. 875--882, 2020.

\end{thebibliography}

\end{document}